\documentclass[journal]{IEEEtran}

\usepackage{cite}

\usepackage{graphicx}

\usepackage{amsmath,amssymb}

\usepackage{array}

\usepackage{url}

\usepackage{booktabs}

\usepackage[caption=false,font=footnotesize]{subfig}

\begin{document}
\title{From Nominal Intensity to Equivalent Rainfall:\\A Path-Based Credibility Evaluation Framework\\for Simulated Rainfall in Autonomous-Driving Perception Tests}

\author{Tian~Xia,
        Xin~Zhao,
        Shaolingfeng~Ye,
        and~Junyi~Chen$^{*}$% <-this % stops a space
\thanks{Tian Xia, Shaolingfeng Ye, and Junyi Chen are with the College of Automotive and Energy Engineering, Tongji University, Shanghai, China.}% <-this % stops a space
\thanks{Xin Zhao is with Tsinghua University, Beijing, China.}% <-this % stops a space
\thanks{$^{*}$Junyi Chen is the corresponding author. E-mail: chenjunyi@tongji.edu.cn.}% <-this % stops a space
\thanks{This work was supported by the National Key Research and Development Program of China under Grant 2024YFB2505702.}}

\maketitle

\begin{abstract}
Credible simulated-rainfall conditions are essential for identifying perception-system performance boundaries and supporting SOTIF-oriented risk assessment in automated driving. However, closed-field simulated-rainfall tests are still commonly described by nominal rainfall intensity or single-point measurements, making it difficult to spatially align simulated rain fields with real rainfall and to reliably map test results to real-world scenarios. To address this problem, this paper proposes a path-based credibility evaluation method for simulated rainfall in autonomous-driving perception tests. Using the drop size--velocity joint distribution of real rainfall as the reference, the proposed method represents each candidate test path by a triplet consisting of path-equivalent rainfall intensity, uncertainty band, and path-averaged Realism of Raindrop Distribution (RRD) score. Lidar target point-cloud count and mean reflectivity are further incorporated to introduce a perception-consistency correction, thereby quantifying the proxy capability of a simulated-rainfall path for real-rainfall perception effects. Experiments are conducted using approximately 10,000 real-rainfall raindrop-spectrum samples, 728 RainSense real-rainfall perception samples, and 45 spatial sampling points within a $2.4~\mathrm{m} \times 7.2~\mathrm{m}$ simulated-rainfall area. The results show that spatial non-uniformity remains under the same nominal condition, confirming the necessity of path-based evaluation. The proposed method identifies Path~IV and Path~VI as preferable candidates, with evaluation results of $11.54 \pm 0.31~\mathrm{mm/h}$, $\mathrm{RRD}=0.43$, and $8.28 \pm 0.34~\mathrm{mm/h}$, $\mathrm{RRD}=0.46$, respectively. These two paths show more balanced performance in rainfall-intensity stability, raindrop-spectrum realism, and perception consistency. The proposed method provides a quantitative basis for path selection, condition description, and credible result interpretation in autonomous-driving perception tests under rainfall, and supports the reliable mapping of closed-field simulated-rainfall test results to real-world scenarios.
\end{abstract}


% \fi
\IEEEpeerreviewmaketitle

\section{Introduction}

Rainfall is both an important environmental component of the operational design domain (ODD) of automated driving systems and a typical trigger condition for perception-system safety and reliability. ISO~34502 and ISO~34503 require automated driving tests to specify road, traffic, environmental, and dynamic-object conditions, and identify weather, visibility, and road-surface states as key elements of test scenarios and ODD descriptions \cite{1,2}. From the perspective of safety of the intended functionality (SOTIF), rainfall may cause perception systems to exceed their performance boundaries, leading to missed detections, false detections, localization errors, or incorrect decisions \cite{3}. In addition, ISO/TC~22/SC~32/WG~16 is advancing the standardization of perception-sensor test methods under adverse weather conditions, including rainfall, based on the automotive lidar basic performance test framework specified in ISO/DIS~13228 \cite{ISO_DIS_13228_2026}. Therefore, as a high-exposure and high-risk trigger condition, rainfall requires systematic testing to clarify its influence on environmental perception systems.

Automated-driving validation under rainfall generally combines virtual testing, proving-ground testing, and real-world road testing. Safety evaluation frameworks such as the PEGASUS project and the UNECE New Assessment/Test Method (NATM) also emphasize the combined use of these complementary approaches to balance efficiency, safety, coverage, and realism \cite{4,5}. Real-world road testing provides the highest environmental fidelity, but rainfall is stochastic and cannot be scheduled or reproduced with controlled intensity, spatial distribution, and traffic context; moreover, real-world testing is costly and inefficient for covering rare or safety-critical conditions \cite{40,41}. Virtual testing is efficient, scalable, and controllable, but it is still limited by the gap between virtual and real environments, especially when modeling the complex effects of adverse weather on sensors and road conditions \cite{40,42}. In contrast, proving-ground testing, and particularly closed-field simulated-rainfall testing, can generate controllable and repeatable rainfall scenarios while retaining physical sensor exposure, making it an important approach for investigating perception degradation and ODD boundaries under rainfall-triggered conditions.

Closed test fields have become important platforms for perception testing under rainfall. Facilities such as Mcity, Jtown, K-City, CARISSMA, and PAVIN provide configurable road, traffic, and environmental conditions for automated-driving and intelligent-transportation tests \cite{6,7,8,9,10,11}. Some of these facilities are further equipped with rainfall, fog, lighting, and other environmental systems to reproduce perception and decision-making challenges under adverse weather. For example, Jtown provides rainfall, fog, sunlight, and street-lighting equipment in its specific environment area, with publicly reported rainfall conditions of 30, 50, and 80~mm/h \cite{7}. The CARISSMA rain system can generate adjustable rainfall with a maximum intensity of approximately 104~mm/h \cite{9}. Cerema's PAVIN fog and rain platform can generate fog and rainfall conditions in a controlled space and has long been used for studies on vision perception and onboard sensors \cite{11}. The development of these facilities indicates that simulated rainfall has become an important environmental capability in closed-field automated-driving tests.

Based on simulated-rainfall facilities or self-built experimental platforms, previous studies have verified the significant influence of rainfall on onboard perception systems. For lidar, Kutila et al. \cite{10}, Montalban et al. \cite{12}, and Choe et al. \cite{13} investigated point-cloud count, reflectivity, and target visibility under artificial-rainfall environments with different rainfall intensities, distances, observation angles, and target conditions. Their results show that rainfall can introduce noise points, weaken target echoes, and degrade target point-cloud quality. For cameras, Li et al. \cite{14} and Pao et al. \cite{15} studied the effects of rainfall, illumination, wind-driven rain, and raindrop adhesion on visual sensing and optical propagation. Their findings indicate that rainfall can reduce image clarity and alter target edges, brightness, and contrast, thereby affecting visual detection performance. For radar, Weihmayr et al. \cite{16}, Gourova et al. \cite{17}, and Slavik and Mishra \cite{18} investigated rainfall effects from the perspectives of multi-sensor comparison, rain-clutter detection using 77~GHz radar, and millimeter-wave propagation attenuation, respectively. Their findings suggest that radar is generally more robust to rainfall than camera and lidar, but it can still be affected by raindrop scattering, clutter, and propagation loss. Overall, rainfall affects onboard sensors through mechanisms such as scattering, attenuation, occlusion, surface adhesion, and propagation loss, thereby influencing object detection and environmental perception results.

However, existing simulated-rainfall tests still lack unified specifications and are mostly described using device settings, nominal rainfall intensity, or meteorological rainfall levels. Although such descriptions are intuitive, they cannot fully characterize the rainfall state within the test space or adequately meet the needs of sensor testing. On the one hand, rainfall-level settings vary across test fields, devices, and studies. On the other hand, even under the same nominal rainfall intensity, the actual rainfall intensity, raindrop spectrum, spatial uniformity, and sensor-exposure conditions may differ. Calibration results from the ROADVIEW project on the CARISSMA outdoor synthetic rain field and the Cerema PAVIN fog and rain platform show that clear deviations can exist between platform-set rainfall intensity and measured rainfall intensity, and that the spatial distribution of the rain field can also be non-uniform \cite{19}. This means that using only nominal rainfall intensity to describe simulated-rainfall conditions may overestimate the comparability and credibility of test conditions.

Therefore, simulated-rainfall evaluation for autonomous-driving perception tests needs to move beyond nominal rainfall intensity and toward the credibility assessment of test conditions. For perception-system testing, rainfall does not merely serve as an environmental background; it directly affects sensor input quality and downstream perception results. Accordingly, simulated-rainfall conditions should be evaluated not only for controllability and repeatability, but also for their ability to represent real rainfall and support condition design, test comparison, and credible interpretation.

To meet the need for credible rainfall representation in autonomous-driving perception tests, this paper proposes a path-based credibility evaluation method for simulated rainfall. Rather than treating nominal intensity or isolated measurement points as sufficient descriptors of the test condition, the proposed method takes the perception test path as the evaluation object and examines whether the rainfall exposure experienced along the path can serve as a credible proxy for real rainfall. Specifically, the evaluation is organized from three complementary aspects: physical realism, which compares the drop size--velocity distribution with real-rainfall references; equivalent rainfall and stability, which determines the real-rainfall intensity represented by the path and the uncertainty of this interpretation; and perception consistency, which checks whether the sensor response under simulated rainfall remains consistent with real-rainfall observations. The main contributions of this paper are summarized as follows:

\begin{itemize}
    \item A path-based credibility evaluation framework for simulated rainfall is proposed for autonomous-driving perception tests. The framework represents each candidate test path using a unified triplet of path-equivalent rainfall intensity, uncertainty band, and path realism score, thereby providing a quantitative basis for test-path selection, condition description, and perception-experiment design.

    \item A high-density spatial sampling analysis of the simulated rain field is conducted. Local rainfall intensity and drop size--velocity joint distribution data are collected within the simulated-rainfall test area to reveal the spatial intensity distribution, microphysical distribution patterns, and their differences from real rainfall.
\end{itemize}

\begin{figure*}[!t]
    \centering
    \includegraphics[width=\textwidth]{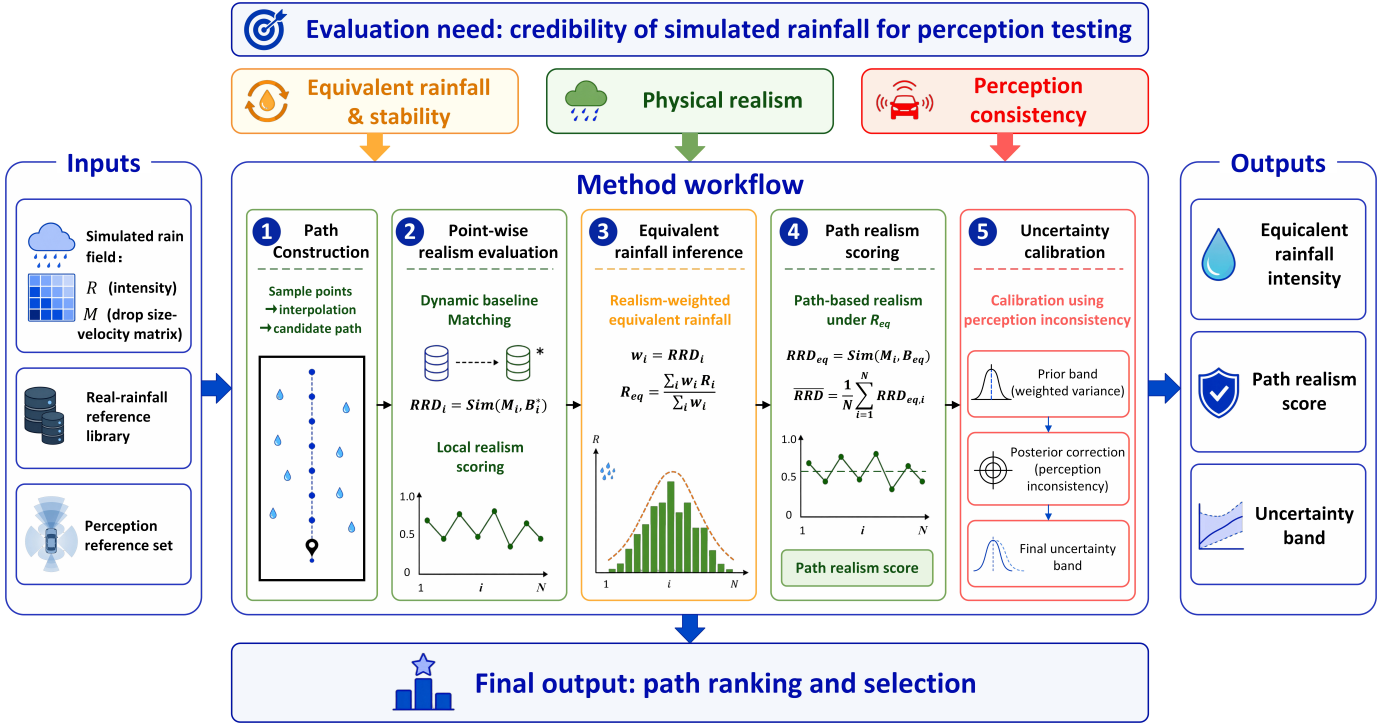}
    \caption{Overall framework of the proposed path-based credibility evaluation method for simulated rainfall in autonomous-driving perception tests. The framework takes the simulated rain field, real-rainfall reference library, and perception reference set as inputs, and outputs the path-equivalent rainfall intensity, path realism score, and prior/posterior uncertainty band for path ranking and selection.}
    \label{fig:framework}
\end{figure*}

\section{Related Work}

\subsection{Perception-Degradation Tests under Simulated Rainfall}

Existing perception tests under rainfall commonly define simulated-rainfall conditions by nominal rainfall intensity or rainfall levels, which are typically controlled through spray flow rate, nozzle combinations, or the number of rain layers, and are then used to evaluate onboard sensors such as lidar, camera, and radar. For example, Hasirlioglu et al. developed a layer-controllable rain simulator with an average rainfall intensity range of 12--120~mm/h and measured the drop size distribution of simulated rainfall using a laser precipitation monitor \cite{20}. Kutila et al. conducted automotive lidar tests using the Cerema fog and rain platform, where rainfall intensity can be controlled and measured with rain gauges and raindrop-spectrum instruments \cite{10}. Weihmayr et al. investigated the responses of camera, lidar, and radar under controlled weather and illumination conditions, including rainfall intensities of 16~mm/h and 98~mm/h \cite{16}. Li et al. further used 25~mm/h and 100~mm/h rainfall conditions in an outdoor rainfall and lighting simulation facility to represent moderate and heavy rain levels and analyze sensor detection performance \cite{14}. These studies show that rainfall intensity is a convenient and widely used control variable in simulated-rainfall tests. However, from the perspective of rain-field physics, rainfall intensity is only a macroscopic descriptor and cannot capture important factors such as raindrop-spectrum structure, spatial distribution, and local fluctuations.

Rain-field calibration studies further show that nominal rainfall intensity may deviate from the actual rain-field state. The ROADVIEW project measured the CARISSMA outdoor synthetic rain field and the Cerema PAVIN fog and rain platform, and reported that nominal rainfall intensities of 10, 25, and 50~mm/h in CARISSMA corresponded to measured mean values of approximately 8.7, 17.4, and 31.2~mm/h, respectively. In PAVIN, target settings of 17, 50, 101, and 175~mm/h corresponded to measured rainfall intensities of approximately 19, 39, 83, and 139~mm/h, with relative deviations of about 12\%--26\% between target and measured values \cite{19}. Similarly, Rasshofer et al. pointed out that indoor rain simulators can be used for sensor evaluation, but their spatial distributions are not necessarily uniform and rain-density peaks may appear below the nozzles \cite{21}. Montalban et al. also noted that rainfall intensity in a climatic chamber was not directly measured but estimated from flow rate and spatial area, while different nozzle configurations could change the drop size, spatial distribution, and spray pattern \cite{12}. These findings indicate that the key issue in simulated-rainfall testing is not only the nominal intensity value itself, but whether the corresponding facility setting can generate an actual rain field consistent with the target rainfall state.

Such deviations in the rain-field state can be further transferred to perception results, thereby affecting the credibility and comparability of test conclusions. Taking lidar point-cloud indicators as an example, most studies report that increasing rainfall weakens target echoes and reduces effective target points or reflectivity due to raindrop scattering, occlusion, and absorption \cite{22,23}. However, some studies reveal more complex behaviors. Montalban et al. observed non-monotonic degradation under 20--120~mm/h artificial rainfall: target point counts decreased at medium-to-high intensities but partially recovered at 120~mm/h, which was attributed to changes in nozzle configuration, drop size distribution, spatial distribution, and spray pattern across nominal rainfall levels \cite{12}. Choe et al. further found that echo intensity decreased with rainfall intensity, distance, and observation angle, whereas the number of point clouds was not sensitive to rainfall intensity and was mainly affected by geometric factors such as distance, observation angle, and effective scanning area \cite{13}. These inconsistent trends indicate that nominal rainfall intensity alone is insufficient to explain lidar point-cloud variations without characterizing the raindrop spectrum, spatial distribution, and local rain-field state.

\begin{table*}[!t]
\centering
\caption{Degradation Patterns of Lidar Perception under Physically Simulated Rainfall}
\label{tab:lidar_degradation}
\renewcommand{\arraystretch}{1.15}
\setlength{\tabcolsep}{4pt}
\begin{tabular*}{\textwidth}{@{\extracolsep{\fill}}l l l c c l@{}}
\hline
Reference & Test Condition & Target & Point Count & Echo Intensity & Main Implication \\
\hline
Kim et al. \cite{22} 
& KICT; 0--50~mm/h 
& Road materials 
& $\downarrow$ 
& $\downarrow$ 
& Monotonic degradation \\

Heinzler et al. \cite{23} 
& Cerema; 55~mm/h 
& Vehicle/pedestrian 
& $\downarrow$ 
& $\downarrow$ 
& Degradation under rainfall \\

Montalban et al. \cite{12} 
& Cerema; 20--120~mm/h 
& Reflective target 
& $\downarrow\uparrow$ 
& $\downarrow\uparrow$ 
& Non-monotonic degradation \\

Choe et al. \cite{13} 
& Darkroom; 20--40~mm/h 
& Target/sign 
& $\sim$ 
& $\downarrow$ 
& NPC insensitive to rain intensity \\
\hline
\multicolumn{6}{@{}l}{\footnotesize Note: $\downarrow$ indicates a decreasing trend, $\sim$ indicates no clear sensitivity, and $\downarrow\uparrow$ indicates non-monotonic variation.}
\end{tabular*}
\end{table*}

\subsection{Indicators for Simulated-Rainfall Evaluation}

Since nominal rainfall intensity alone cannot ensure the credibility and usability of simulated-rainfall conditions, additional indicators are needed to describe and evaluate the rain field. Although most existing studies on simulated-rainfall evaluation originate from hydrology, soil and water conservation, and soil erosion research, their indicator systems for raindrop microphysical characteristics provide useful references for autonomous-driving perception tests.

Existing indicators can be broadly divided into three categories: macroscopic statistical indicators, microphysical distribution indicators, and physical-effect indicators. Macroscopic statistical indicators include rainfall intensity, the coefficient of variation of rainfall intensity ($CV$), the Christiansen uniformity coefficient ($CU$), and low-quarter distribution uniformity ($DU_{lq}$), which describe the rainfall level, intensity fluctuation, and spatial uniformity of a simulated rain field \cite{24,27,28}. Among them, $CV$, $CU$, and $DU_{lq}$ have relatively clear value directions and are therefore commonly used to evaluate rainfall stability, spatial uniformity, and low-intensity coverage. However, $CU$ is a single-percentage indicator and may hide local spatial differences, making it sensitive to sampling design \cite{24,26}. Microphysical distribution indicators, such as drop size distribution (DSD), median volumetric drop diameter ($D_{50}$), drop fall velocity, and the drop size--velocity joint distribution matrix, describe raindrop size, velocity, and number characteristics \cite{25,29,30}. Physical-effect indicators, including kinetic energy expenditure ($KE$), kinetic energy per unit rainfall depth ($KE_{\mathrm{vol}}$), and rainfall momentum ($M$), characterize the impact of raindrops on the tested object or ground surface \cite{25,29,31}.

This distinction is important because not all indicators are evaluative by themselves. Some macroscopic uniformity indicators directly imply a preferred direction, whereas most microphysical and physical-effect indicators are primarily descriptive. They characterize specific rain-field properties, but do not indicate whether a simulated rain field is realistic unless a reference is introduced. Therefore, these descriptive indicators can be transformed into evaluation criteria only when compared with real-rainfall references, natural terminal-velocity relationships, or natural-rainfall energy ranges at comparable intensities \cite{25,31,32}.

\begin{figure}[!t]
\centering
\includegraphics[width=0.9\linewidth]{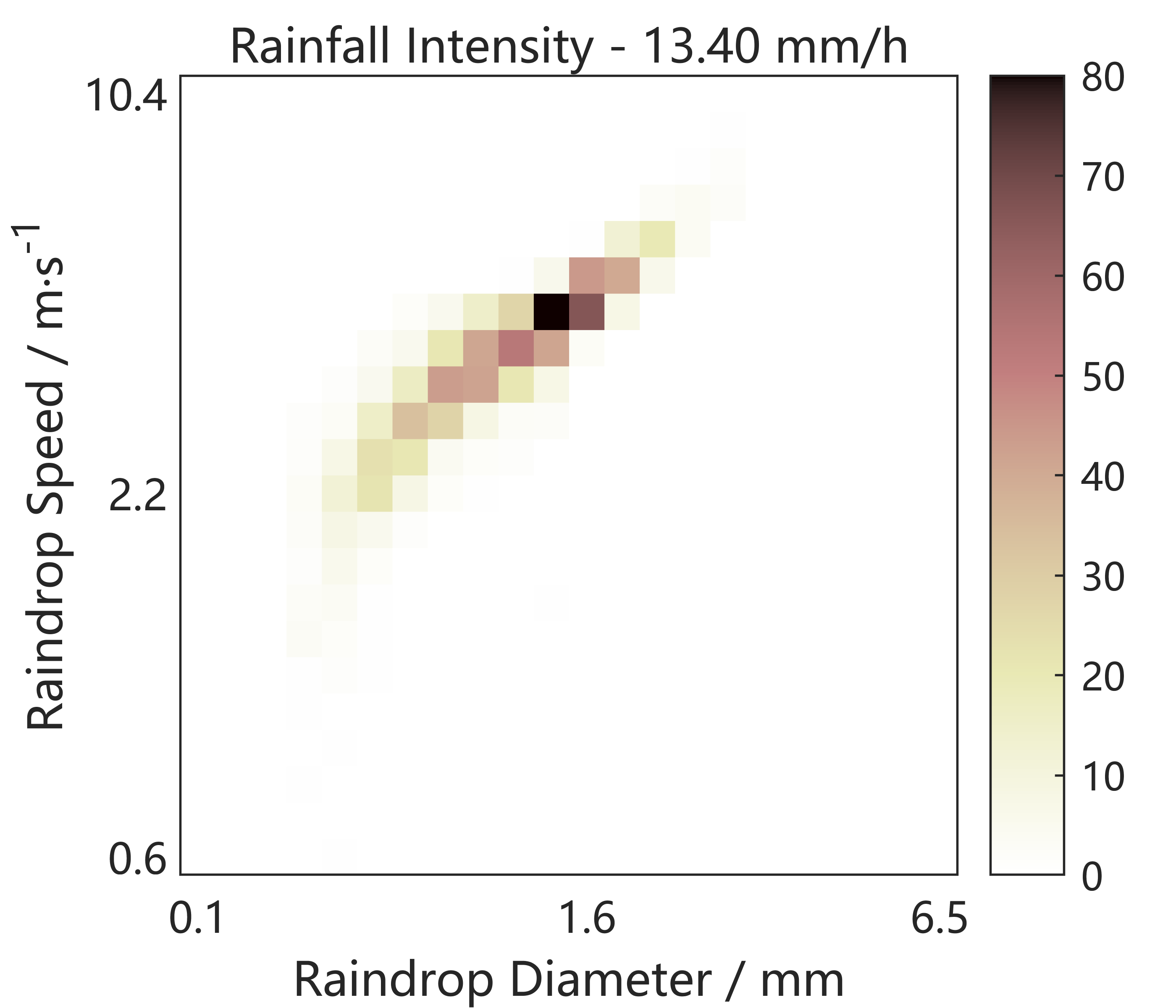}
\caption{Schematic illustration of the drop size--velocity joint distribution.}
\label{fig:dsd_velocity_matrix}
\end{figure}

For autonomous-driving perception tests, simulated-rainfall evaluation should shift from ground-surface impact processes to sensor-exposure processes. Macroscopic statistical indicators describe the overall intensity and uniformity of a rain field, but they cannot capture the local raindrop structures encountered by a sensor along a test path. Physical-effect indicators are more suitable for soil erosion and water-conservation studies because they focus on raindrop impact, splash erosion, and erosivity on the ground surface \cite{26,28}. By contrast, microphysical distribution indicators describe fundamental raindrop information, including size, velocity, and number, and are more closely related to the physical processes through which rainfall affects lidar echoes, camera imaging, and radar scattering \cite{12,15,16,18}. In particular, as shown in Fig.~\ref{fig:dsd_velocity_matrix}, the drop size--velocity joint distribution preserves raindrop size, velocity, and number simultaneously, and can further support the derivation of rainfall intensity, raindrop-spectrum characteristics, kinetic energy, and momentum. Therefore, using the drop size--velocity joint distribution as the core rain-field representation is suitable for comparing microphysical differences between simulated rainfall and real rainfall, and provides a basis for subsequent path-based realism evaluation and perception-consistency analysis.

\subsection{Credibility Evaluation of Simulated Test Conditions}

Credibility evaluation under real-world references aims to quantify the gap between simulated conditions and the real-world environment, and to determine whether this gap is acceptable for the intended task. In automated driving, the credible use of simulated conditions, whether in physical simulation, sensor simulation, or data augmentation, cannot rely only on controllability and repeatability, but also requires validation against real-world data. Daza et al. pointed out that the validity of automated-driving simulation depends on whether the reality gap is sufficiently small for the target task \cite{33}. Existing studies can be broadly divided into two categories: direct evaluation, which measures the physical or statistical similarity between simulated and real conditions, and indirect evaluation, which examines whether task outputs under simulated conditions are consistent with real-world system behavior.

Direct evaluation assesses the realism of simulated conditions using statistics or feature distributions derived from real data. For example, Neural NDE proposed statistical realism, requiring generated naturalistic driving environments to match naturalistic driving data not only in safety-critical scenario coverage, but also in distributions of crash rate, crash type, crash severity, near-miss events, vehicle speed, following distance, and yielding behavior \cite{34}. Song et al. defined the realism of critical scenarios as the similarity between synthetic and real critical scenarios, measured through scenario-attribute distributions and distances in a vectorized scenario space \cite{35}. Synthetic-data studies also quantify the synthetic-to-real gap using feature embeddings, image-style distributions, or scenario attributes \cite{36}. These studies indicate that simulated-condition realism should be established under real-world references and measured through computable distributional differences. For simulated-rainfall testing, the corresponding question is whether the simulated rain field is close to real rainfall in rainfall intensity, drop size--velocity distribution, and spatial distribution, rather than being defined only by device settings.

Indirect evaluation verifies the real-world proxy capability of simulated conditions through downstream task performance or system-output consistency. For driving tasks, the key question is whether simulation can reproduce motion responses and behavioral changes observed in real-vehicle or real-road experiments. For example, Daza et al. evaluated the real-to-sim gap in a path-tracking task by comparing steering angle, lateral error, lateral acceleration, jerk, and other output signals from the same planning and control system in CARLA and real-road tests, using Pearson correlation and maximum normalized cross-correlation \cite{33}. For perception tasks, the focus shifts to whether simulated perception data can reproduce recognition performance or degradation patterns observed in real data. Virtual KITTI compares multi-object tracking performance between cloned virtual sequences and real KITTI sequences \cite{37}; RadSimReal validates object detection using physics-based synthetic radar data and real radar data \cite{38}; and S2R-Bench evaluates perception robustness across real sensor anomalies, normal data, and simulated anomaly conditions \cite{39}. These studies show that input-level realism alone is insufficient, and task-related indicators are still needed to verify whether simulated conditions can support reliable test conclusions.

Overall, the credible use of simulated conditions needs to answer two questions: whether the gap between simulated and real conditions is measurable, and whether this gap affects downstream test conclusions. For simulated-rainfall testing, the evaluation focus should therefore shift from device-set nominal rainfall intensity to rain-field difference analysis under real-rainfall references, and further examine whether perception indicators under simulated rainfall follow the patterns observed under real rainfall. Accordingly, simulated-rainfall evaluation for autonomous-driving perception tests should consider both rain-field realism and perception consistency, rather than relying solely on nominal rainfall intensity to judge whether a test condition is usable.

\section{Problem Formulation and Evaluation Framework}

\subsection{Problem Formulation}

\begin{figure}[!t]
    \centering
    \includegraphics[width=0.95\linewidth]{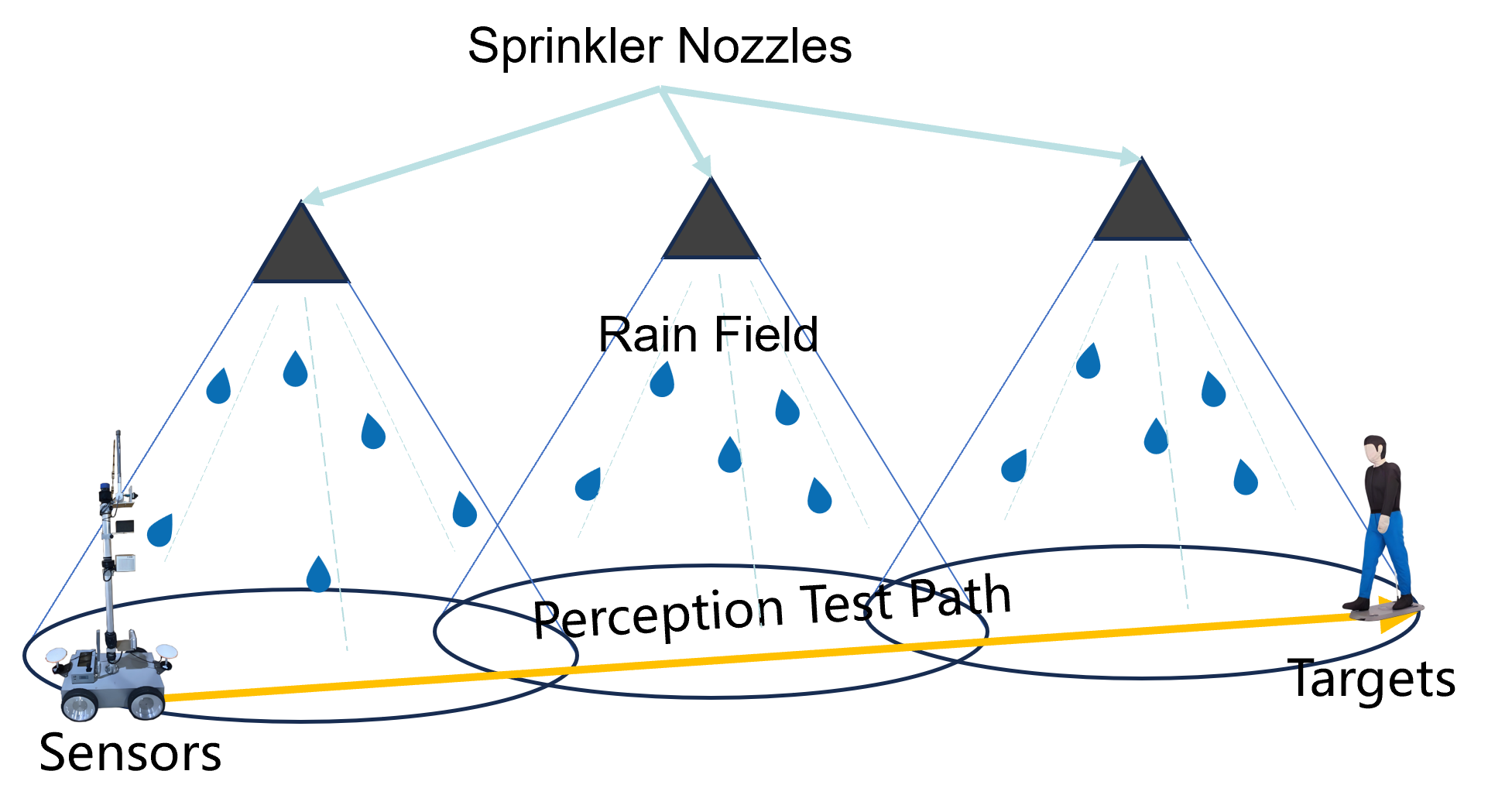}
    \caption{Schematic illustration of the perception test path under a simulated rain field.}
    \label{fig:perception_path}
\end{figure}

In proving-ground tests of autonomous-driving perception systems, simulated rainfall is commonly used to reproduce the effects of real rainfall on sensors such as camera, lidar, and radar. However, as illustrated in Fig.~\ref{fig:perception_path}, the rainfall condition experienced by the perception system under test cannot be sufficiently described by the rainfall intensity measured at a fixed point. Instead, it corresponds to a sequence of spatially varying rainfall states encountered along a specific perception test path. The preceding review indicates that rainfall intensity only describes the macroscopic level of a rain field and cannot capture its instantaneous microphysical state. In contrast, microphysical distributions are closely related to perception-degradation mechanisms, such as lidar echo attenuation, camera imaging degradation, and radar scattering. Therefore, for the perception system under test, the rainfall condition should be represented as a sequence of local rainfall states along the test path, where each local state is jointly characterized by local rainfall intensity and the drop size--velocity distribution.

Based on this understanding, this paper defines the evaluation object of simulated-rainfall conditions as the path-based rainfall state. For a candidate perception test path, different locations along the path may have different local rainfall intensities and microphysical rain-field characteristics. The objective of path-based evaluation is not to determine whether a single measurement point is close to real rainfall, but to assess whether the rainfall exposure process formed by the entire path can represent a certain class of real-rainfall conditions and support credible interpretation of perception test results. Accordingly, the credibility evaluation of simulated rainfall should jointly consider the microphysical realism of local raindrop structures along the path, the stability of the path-equivalent rainfall intensity, and the consistency between perception indicators under the path and real-rainfall references.

\subsection{Evaluation Objectives for Simulated-Rainfall Credibility}

The credibility of simulated rainfall discussed in this paper does not refer to whether the rainfall facility operates normally, nor does it simply indicate whether a nominal rainfall intensity can be achieved. Instead, it concerns whether a simulated-rainfall path can represent a certain class of real-rainfall conditions in autonomous-driving perception tests. For this purpose, the credibility of a simulated-rainfall path should be evaluated from at least three aspects.

First, simulated rainfall should be physically close to real rainfall. The drop size--velocity distribution along the path should be similar to that of real rainfall under the same or comparable rainfall intensity. If a simulated-rainfall condition reaches a given intensity level but its raindrop-spectrum structure differs substantially from real rainfall, its physical realism is insufficient.

Second, the simulated-rainfall path should provide a clear interpretation of equivalent rainfall intensity. Since the perception system observes a series of local rainfall states along the path, the path condition cannot be represented by the measurement result at a single point. The key question is which real-rainfall intensity the entire path is closest to, and whether this path-equivalent intensity is stable. If the rainfall states within the path vary strongly, or if high-realism regions correspond to a dispersed intensity distribution, an equivalent rainfall intensity can still be estimated, but its credibility should be accompanied by an uncertainty description.

Third, simulated rainfall should be consistent with real rainfall at the perception-response level. In autonomous-driving perception tests, rainfall ultimately affects sensors and perception algorithms. Therefore, even if a path is close to real rainfall in terms of raindrop-spectrum structure, it is still necessary to examine whether the resulting perception indicators fall within a reasonable range observed under real-rainfall conditions. If the perception response under simulated rainfall deviates significantly from the real-rainfall reference distribution, the confidence in the corresponding path-based test result should be reduced.

\subsection{Path-Based Credibility Evaluation Framework}

To address the above objectives, this paper proposes a path-based credibility evaluation framework for simulated rainfall in proving-ground tests of autonomous-driving perception systems, as shown in Fig.~\ref{fig:framework}. The framework takes the path-based rainfall state as the evaluation object and decomposes the credibility of a simulated-rainfall path into three interrelated questions: whether the local rainfall states along the path have microphysical raindrop distributions close to real rainfall; which real-rainfall intensity the entire path can equivalently represent, and whether this equivalent-intensity interpretation is stable; and whether the perception response induced by simulated rainfall falls within the reasonable range observed under real-rainfall conditions.

Following this idea, the proposed method first converts discrete measurement-point data in the simulated rain field into a path-based rainfall-state representation. A real-rainfall baseline matched with the local rainfall intensity is then constructed at the point-wise level to evaluate the rainfall realism of each path point. Next, the point-wise realism scores are used as weights to infer the path-equivalent rainfall intensity, and a basic uncertainty band is estimated according to the intensity dispersion within the path. The uncertainty band is further corrected by considering insufficient path realism and perception-consistency deviations. Finally, the mean path realism is re-evaluated under the path-equivalent rainfall intensity, forming a comprehensive evaluation result for candidate path ranking and selection.

\section{Methodology}

To convert the above credibility evaluation framework into a computable basis for path selection, this section defines the main computational steps. For the $p$-th candidate test path, the proposed method sequentially calculates point-wise rainfall realism, path-equivalent rainfall intensity, final uncertainty band, and path realism score. The evaluation result is represented by the following triplet:
\begin{equation}
C_p =
\left(
\hat{R}_{\mathrm{eq},p},
\Delta_{\mathrm{final},p},
\overline{\mathrm{RRD}}_p
\right),
\label{eq:credibility_triplet}
\end{equation}
where $C_p$ denotes the credibility descriptor of the $p$-th path, $\hat{R}_{\mathrm{eq},p}$ is the path-equivalent rainfall intensity, $\Delta_{\mathrm{final},p}$ is the final uncertainty band, and $\overline{\mathrm{RRD}}_p$ is the path realism score. Specifically, $\hat{R}_{\mathrm{eq},p}$ describes the real-rainfall intensity represented by the path, $\Delta_{\mathrm{final},p}$ characterizes the uncertainty of this intensity interpretation, and $\overline{\mathrm{RRD}}_p$ evaluates the microphysical realism of the path under the corresponding equivalent rainfall intensity.

In candidate-path comparison, the triplet $C_p$ is not intended to optimize a single indicator independently, but to support path selection under a target rainfall intensity $R_{\mathrm{tar}}$. Specifically, $\hat{R}_{\mathrm{eq},p}$ should be close to $R_{\mathrm{tar}}$ to ensure rainfall-intensity compatibility, $\Delta_{\mathrm{final},p}$ should be small to ensure a stable equivalent-intensity interpretation, and $\overline{\mathrm{RRD}}_p$ should be large to ensure high microphysical realism. Therefore, candidate-path ranking is formulated as a joint comparison of rainfall-intensity compatibility, uncertainty, and realism, rather than a binary decision based on fixed thresholds.

\subsection{Path Construction}

Assume that a simulated rain field contains a set of spatial sampling points. At each sampling point, the local rainfall intensity and the drop size--velocity distribution matrix can be obtained. The drop size--velocity distribution matrix describes the number of raindrops in different diameter and velocity bins, and reflects the microphysical structure of simulated rainfall.

For the $p$-th candidate test path, the path is discretized into $N_p$ ordered evaluation points:
\begin{equation}
P_p =
\left\{
\left(R_{p,i}, M_{p,i}\right)
\right\}_{i=1}^{N_p},
\label{eq:path_definition}
\end{equation}
where $R_{p,i}$ denotes the local rainfall intensity at the $i$-th evaluation point of the $p$-th path, $M_{p,i}\in\mathbb{R}^{32\times32}$ denotes the corresponding drop size--velocity distribution matrix, and $N_p$ is the number of evaluation points on the path.

In practical test fields, instruments such as Laser-Optical Disdrometers can only acquire rainfall data at limited sampling points. Therefore, this paper first constructs the spatial rain field based on the rainfall intensity and raindrop-spectrum matrix measured at sampling points, and then obtains $R_{p,i}$ and $M_{p,i}$ at each evaluation point through inter-path interpolation. This process converts the evaluation object from discrete sampling points to a continuous rainfall state along the actual driving path, and also enables the estimation of intermediate paths based on measurements from adjacent paths.

\subsection{Point-Wise Realism Evaluation}

The calculation of path-equivalent rainfall intensity requires determining whether the local rainfall state at each path point is close to real rainfall. Therefore, for each path point, a point-wise real-rainfall baseline is constructed using real-rainfall samples with rainfall intensities close to the local simulated rainfall intensity. The similarity between the simulated rainfall matrix and this baseline is then used to evaluate point-wise rainfall realism. It should be emphasized that the baseline is constructed separately for each point; that is, the baseline for the $i$-th point of the $p$-th path is determined by its local rainfall intensity $R_{p,i}$.

Let the real-rainfall reference library be
\begin{equation}
\mathcal{D}_{\mathrm{real}} =
\left\{
\left(R_j^{\mathrm{real}}, M_j^{\mathrm{real}}\right)
\right\}_{j=1}^{N_{\mathrm{real}}},
\label{eq:real_reference_library}
\end{equation}
where $R_j^{\mathrm{real}}$ denotes the rainfall intensity of the $j$-th real-rainfall sample, and $M_j^{\mathrm{real}}$ denotes the corresponding real-rainfall drop size--velocity distribution matrix.

For path point $(p,i)$, real-rainfall samples with rainfall intensities close to $R_{p,i}$ are selected from the reference library to form the effective point-wise sample set:
\begin{equation}
\mathcal{S}_{p,i}^{\mathrm{pt}} =
\left\{
j \mid
\left| R_j^{\mathrm{real}} - R_{p,i} \right|
\leq
\tau \left(R_{p,i}\right)
\right\},
\label{eq:pointwise_sample_set}
\end{equation}
where $\tau(R_{p,i})$ is the rainfall-intensity tolerance band, determined by both an absolute tolerance and a relative tolerance:
\begin{equation}
\tau \left(R_{p,i}\right)
=
\max \left(\tau_0, \eta R_{p,i}\right),
\label{eq:tolerance_band}
\end{equation}
where $\tau_0$ is the minimum rainfall-intensity tolerance and $\eta$ is the relative tolerance coefficient. In this study, both parameters are set to $0.1$.

When the effective sample set contains many samples, the samples are sorted in ascending order according to $\left|R_j^{\mathrm{real}}-R_{p,i}\right|$, and the nearest $K$ samples are selected as the point-wise evaluation samples. The point-wise real-rainfall baseline is then obtained by averaging their drop size--velocity matrices:
\begin{equation}
B_{p,i}^{\mathrm{pt}}
=
\frac{1}{K}
\sum_{j\in \mathcal{S}_{p,i,K}^{\mathrm{pt}}}
M_j^{\mathrm{real}},
\label{eq:pointwise_baseline}
\end{equation}
where $\mathcal{S}_{p,i,K}^{\mathrm{pt}}$ denotes the selected $K$ effective real-rainfall samples. This averaging process reduces the randomness of individual real-rainfall samples and improves the stability of the point-wise baseline.

After obtaining $B_{p,i}^{\mathrm{pt}}$, the structural similarity between the simulated rainfall matrix $M_{p,i}$ and the real-rainfall baseline matrix $B_{p,i}^{\mathrm{pt}}$ is calculated. This paper uses the structural similarity index to measure the closeness of their overall distribution patterns:
\begin{equation}
\mathrm{SSIM}
\left(
M_{p,i}, B_{p,i}^{\mathrm{pt}}
\right)
=
\frac{
\left(2\mu_M\mu_B+C_1\right)
\left(2\sigma_{MB}+C_2\right)
}{
\left(\mu_M^2+\mu_B^2+C_1\right)
\left(\sigma_M^2+\sigma_B^2+C_2\right)
},
\label{eq:ssim}
\end{equation}
where $\mu_M$ and $\mu_B$ are the element-wise means of $M_{p,i}$ and $B_{p,i}^{\mathrm{pt}}$, respectively; $\sigma_M^2$ and $\sigma_B^2$ are the corresponding element-wise variances; $\sigma_{MB}$ is the covariance between the two matrices; and $C_1$ and $C_2$ are constants used to avoid division by zero.

Using structural similarity alone may ignore local anomalies. For example, a simulated rainfall matrix may contain abnormal drop size--velocity combinations that do not appear in the real-rainfall baseline, or some matrix elements may differ from the baseline by a large magnitude. Therefore, a point-wise penalty coefficient is introduced to correct the structural similarity result.

The abnormal distribution ratio is defined as
\begin{equation}
e_{p,i}^{\mathrm{dist}}
=
\frac{
N_{p,i}^{\mathrm{dist}}
}{
N_{p,i}^{\mathrm{nz}}
},
\label{eq:distribution_anomaly}
\end{equation}
where $N_{p,i}^{\mathrm{dist}}$ denotes the number of abnormal elements whose values are nonzero in $M_{p,i}$ but zero at the corresponding positions in $B_{p,i}^{\mathrm{pt}}$, and $N_{p,i}^{\mathrm{nz}}$ denotes the number of nonzero elements in $M_{p,i}$.

The abnormal magnitude ratio is further defined as
\begin{equation}
e_{p,i}^{\mathrm{mag}}
=
\frac{
N_{p,i}^{\mathrm{mag}}
}{
N_{p,i}^{\mathrm{nz}}
},
\label{eq:magnitude_anomaly}
\end{equation}
where $N_{p,i}^{\mathrm{mag}}$ denotes the number of corresponding nonzero elements in which the simulated rainfall matrix and the real-rainfall baseline matrix differ by a large magnitude.

Combining the two types of anomalies, the point-wise penalty coefficient is defined as
\begin{equation}
\phi_{p,i}^{\mathrm{pt}}
=
1
-
\lambda_1 e_{p,i}^{\mathrm{dist}}
-
\lambda_2 e_{p,i}^{\mathrm{mag}},
\label{eq:pointwise_penalty_raw}
\end{equation}
where $\lambda_1$ and $\lambda_2$ are the penalty weights for abnormal distribution and abnormal magnitude, respectively. To avoid negative penalty coefficients, truncation is applied:
\begin{equation}
\phi_{p,i}^{\mathrm{pt}}
=
\max
\left(
\phi_{\min},
\phi_{p,i}^{\mathrm{pt}}
\right).
\label{eq:pointwise_penalty}
\end{equation}

Finally, the point-wise rainfall realism of the $i$-th point on the $p$-th path is defined as
\begin{equation}
\mathrm{RRD}_{p,i}^{\mathrm{pt}}
=
\left[
\mathrm{SSIM}
\left(
M_{p,i}, B_{p,i}^{\mathrm{pt}}
\right)
\right]^{\xi}
\cdot
\phi_{p,i}^{\mathrm{pt}},
\label{eq:pointwise_rrd}
\end{equation}
where $\xi$ is the similarity gain coefficient, used to adjust the influence of structural similarity on the final realism score. A higher $\mathrm{RRD}_{p,i}^{\mathrm{pt}}$ indicates that the simulated raindrop-spectrum structure at this path point is closer to real rainfall with a matched rainfall intensity.

\subsection{Path-Equivalent Rainfall Intensity and Stability Evaluation}

\subsubsection{Path-Equivalent Rainfall Intensity}

After obtaining the point-wise realism scores along the path, they are used as weights for rainfall-intensity aggregation. For the $i$-th point of the $p$-th path, the rainfall-intensity weight is defined as
\begin{equation}
w_{p,i}
=
\mathrm{RRD}_{p,i}^{\mathrm{pt}}.
\label{eq:weight_linear}
\end{equation}
A nonlinear enhancement form can also be introduced if needed:
\begin{equation}
w_{p,i}
=
\left(
\mathrm{RRD}_{p,i}^{\mathrm{pt}}
\right)^{\gamma},
\label{eq:weight_nonlinear}
\end{equation}
where $\gamma$ is the weight adjustment coefficient. When $\gamma=1$, the weight equals the point-wise realism score, which is used in this study. When $\gamma>1$, high-realism points are further emphasized, whereas low-realism points contribute less.

Based on the above weights, the path-equivalent rainfall intensity of the $p$-th path is defined as
\begin{equation}
\hat{R}_{\mathrm{eq},p}
=
\frac{
\sum_{i=1}^{N_p}
w_{p,i} R_{p,i}
}{
\sum_{i=1}^{N_p}
w_{p,i}
}.
\label{eq:path_equivalent_intensity}
\end{equation}
This value represents the best representative rainfall intensity of the path after jointly considering local rainfall intensity and local realism. Compared with a simple arithmetic mean, $\hat{R}_{\mathrm{eq},p}$ reduces the influence of low-realism path points and emphasizes points whose microphysical raindrop-spectrum structures are closer to real rainfall. Therefore, $\hat{R}_{\mathrm{eq},p}$ can be regarded as a quality-corrected representative rainfall intensity, and serves as the basis for subsequent path selection and path realism evaluation.

\subsubsection{Basic Uncertainty Band}

A point estimate of path-equivalent rainfall intensity is insufficient to describe the reliability of a path condition. If local rainfall intensities vary strongly along the path, or if high-weight points correspond to a dispersed intensity distribution, the path may still have a weak representative capability even though an equivalent intensity can be calculated. Therefore, a basic uncertainty band is further estimated.

The weighted variance of rainfall intensity along the path is calculated as
\begin{equation}
\sigma_p^2
=
\frac{
\sum_{i=1}^{N_p}
w_{p,i}
\left(
R_{p,i}
-
\hat{R}_{\mathrm{eq},p}
\right)^2
}{
\sum_{i=1}^{N_p}
w_{p,i}
},
\label{eq:weighted_variance}
\end{equation}
where $\sigma_p^2$ reflects the dispersion of local rainfall intensity around $\hat{R}_{\mathrm{eq},p}$.

To avoid overly optimistic uncertainty estimation caused by uneven weight distributions, the effective sample size is introduced:
\begin{equation}
n_{\mathrm{eff},p}
=
\frac{
\left(
\sum_{i=1}^{N_p}
w_{p,i}
\right)^2
}{
\sum_{i=1}^{N_p}
w_{p,i}^2
}.
\label{eq:effective_sample_size}
\end{equation}
The basic standard error of the path-equivalent rainfall intensity is then obtained as
\begin{equation}
\Delta_p
=
SE_p
=
\sqrt{
\frac{
\sigma_p^2
}{
n_{\mathrm{eff},p}
}
}.
\label{eq:basic_uncertainty}
\end{equation}
Thus, the basic expression of the path-equivalent rainfall intensity is
\begin{equation}
\hat{R}_{\mathrm{eq},p}
\pm
\Delta_p.
\label{eq:basic_equivalent_result}
\end{equation}
A smaller $\Delta_p$ indicates that the rainfall intensities corresponding to high-realism points are more concentrated, and that the path can more stably represent real-rainfall conditions near $\hat{R}_{\mathrm{eq},p}$. A larger $\Delta_p$ indicates stronger within-path rainfall fluctuation and greater uncertainty in the equivalent-intensity interpretation.

\subsubsection{Uncertainty-Band Correction Based on Rainfall Realism}

The basic uncertainty band $\Delta_p$ mainly reflects within-path rainfall-intensity fluctuation, but it does not fully express the additional risk caused by insufficient rainfall realism. For example, a path may have a stable intensity distribution while its raindrop-spectrum structure deviates from real rainfall. In this case, using only $\Delta_p$ would underestimate the uncertainty of the path as a test condition.

Therefore, a realism correction coefficient is introduced for the first-stage correction of the uncertainty band. The preliminary path realism is calculated from the point-wise realism scores used for equivalent-intensity inference:
\begin{equation}
\overline{\mathrm{RRD}}_{p}^{\mathrm{pt}}
=
\frac{1}{N_p}
\sum_{i=1}^{N_p}
\mathrm{RRD}_{p,i}^{\mathrm{pt}}.
\label{eq:preliminary_path_rrd}
\end{equation}
Let $\mathrm{RRD}_{\mathrm{thr}}$ denote the realism threshold. The degree of realism insufficiency is defined as
\begin{equation}
d_p^{\mathrm{rrd}}
=
\frac{
\mathrm{RRD}_{\mathrm{thr}}
-
\overline{\mathrm{RRD}}_{p}^{\mathrm{pt}}
}{
\mathrm{RRD}_{\mathrm{thr}}
}.
\label{eq:rrd_deficiency_raw}
\end{equation}
Truncation is then applied:
\begin{equation}
d_p^{\mathrm{rrd}}
=
\min
\left(
1,
\max
\left(
0,
d_p^{\mathrm{rrd}}
\right)
\right).
\label{eq:rrd_deficiency}
\end{equation}
When the preliminary path realism reaches or exceeds the threshold, $d_p^{\mathrm{rrd}}=0$. When it is lower than the threshold, $d_p^{\mathrm{rrd}}$ increases with the degree of realism insufficiency.

The realism correction coefficient is further defined as
\begin{equation}
\phi_p^{\mathrm{rrd}}
=
1
+
\alpha d_p^{\mathrm{rrd}},
\label{eq:rrd_correction_factor}
\end{equation}
where $\alpha$ is the amplification coefficient for realism insufficiency. The corrected uncertainty band is then
\begin{equation}
\Delta_p'
=
\phi_p^{\mathrm{rrd}}
\Delta_p.
\label{eq:rrd_corrected_uncertainty}
\end{equation}
This correction indicates that the basic uncertainty caused by within-path intensity fluctuation is further enlarged when rainfall realism is insufficient. It prevents a path with stable intensity but unrealistic raindrop-spectrum structure from being misidentified as a highly credible test path.

\subsection{Path Realism Score}

After calculating the equivalent rainfall intensity and the corrected uncertainty band, the overall realism of the path is further evaluated under its path-equivalent rainfall intensity. The path realism score answers whether the entire path has realistic raindrop-spectrum characteristics when representing the rainfall condition $\hat{R}_{\mathrm{eq},p}$.

After $\hat{R}_{\mathrm{eq},p}$ is obtained, real-rainfall samples matched with $\hat{R}_{\mathrm{eq},p}$ are reselected from the real-rainfall reference library to construct the path-level equivalent-intensity baseline:
\begin{equation}
\mathcal{S}_{p}^{\mathrm{eq}}
=
\left\{
j \mid
\left|
R_j^{\mathrm{real}}
-
\hat{R}_{\mathrm{eq},p}
\right|
\leq
\tau
\left(
\hat{R}_{\mathrm{eq},p}
\right)
\right\}.
\label{eq:eq_sample_set}
\end{equation}
Similarly, the nearest $K$ samples to $\hat{R}_{\mathrm{eq},p}$ are selected to construct the path-level evaluation baseline:
\begin{equation}
B_{p}^{\mathrm{eq}}
=
\frac{1}{K}
\sum_{j\in \mathcal{S}_{p,K}^{\mathrm{eq}}}
M_j^{\mathrm{real}}.
\label{eq:eq_baseline}
\end{equation}
Then, each drop size--velocity matrix $M_{p,i}$ along the path is compared with $B_p^{\mathrm{eq}}$ to obtain the point-wise realism under the equivalent-intensity baseline:
\begin{equation}
\mathrm{RRD}_{p,i}^{\mathrm{eq}}
=
\left[
\mathrm{SSIM}
\left(
M_{p,i}, B_p^{\mathrm{eq}}
\right)
\right]^{\xi}
\cdot
\phi_{p,i}^{\mathrm{eq}},
\label{eq:eq_pointwise_rrd}
\end{equation}
where $\phi_{p,i}^{\mathrm{eq}}$ is the point-wise penalty coefficient calculated based on the path-level baseline, following the same procedure as in the point-wise evaluation described above.

Finally, the path realism score of the $p$-th path is defined as
\begin{equation}
\overline{\mathrm{RRD}}_p
=
\frac{1}{N_p}
\sum_{i=1}^{N_p}
\mathrm{RRD}_{p,i}^{\mathrm{eq}}.
\label{eq:path_rrd_score}
\end{equation}
This score reflects the raindrop-spectrum realism of the entire path under $\hat{R}_{\mathrm{eq},p}$. A higher $\overline{\mathrm{RRD}}_p$ indicates that the path can better represent real rainfall near $\hat{R}_{\mathrm{eq},p}$, whereas a lower value suggests that the path has a clear microphysical discrepancy from real rainfall even though an equivalent rainfall intensity can be estimated.

\subsection{Uncertainty-Band Correction Based on Perception Consistency}

For autonomous-driving perception tests, simulated rainfall should be close to real rainfall not only in raindrop-spectrum structure, but also in perception response. If the perception indicators induced by simulated rainfall significantly deviate from the reference range observed under real-rainfall conditions with the same experimental setup, the path may not reliably represent the influence of real rainfall on the perception system even if it has a certain level of physical realism. Therefore, a perception-consistency correction coefficient is introduced for posterior uncertainty correction.

Let $y_{p,m}$ denote the observed value of the $m$-th perception indicator under simulated-rainfall path $P_p$. According to the path-equivalent rainfall intensity $\hat{R}_{\mathrm{eq},p}$, the corresponding real-rainfall reference band of this indicator is obtained from the real-rainfall perception reference set:
\begin{equation}
\left[
L_m
\left(
\hat{R}_{\mathrm{eq},p}
\right),
U_m
\left(
\hat{R}_{\mathrm{eq},p}
\right)
\right],
\label{eq:perception_reference_band}
\end{equation}
where $L_m(\hat{R}_{\mathrm{eq},p})$ and $U_m(\hat{R}_{\mathrm{eq},p})$ denote the lower and upper bounds of the $m$-th perception indicator near $\hat{R}_{\mathrm{eq},p}$ under real-rainfall conditions, respectively.

The normalized deviation distance of the $m$-th perception indicator is defined as
\begin{equation}
D_{p,m}
=
\frac{
\max
\left(
0,
L_m
\left(
\hat{R}_{\mathrm{eq},p}
\right)
-
y_{p,m},
y_{p,m}
-
U_m
\left(
\hat{R}_{\mathrm{eq},p}
\right)
\right)
}{
U_m
\left(
\hat{R}_{\mathrm{eq},p}
\right)
-
L_m
\left(
\hat{R}_{\mathrm{eq},p}
\right)
}.
\label{eq:perception_deviation}
\end{equation}
When $y_{p,m}$ lies within the real-rainfall reference band, $D_{p,m}=0$. When $y_{p,m}$ falls outside the reference band, $D_{p,m}$ increases with the degree of deviation.

For multiple perception indicators, the aggregated perception-deviation distance is defined as
\begin{equation}
D_p^{\mathrm{per}}
=
\sum_{m=1}^{M}
\omega_m D_{p,m},
\label{eq:aggregated_perception_deviation}
\end{equation}
where $\omega_m$ is the weight of the $m$-th perception indicator and satisfies
\begin{equation}
\sum_{m=1}^{M}
\omega_m
=
1.
\label{eq:perception_weight_constraint}
\end{equation}
The perception-consistency correction coefficient is then defined as
\begin{equation}
\phi_p^{\mathrm{per}}
=
1
+
\beta D_p^{\mathrm{per}},
\label{eq:perception_correction_factor}
\end{equation}
where $\beta$ is the amplification coefficient for perception deviation. The final uncertainty band is
\begin{equation}
\Delta_{\mathrm{final},p}
=
\phi_p^{\mathrm{per}}
\Delta_p'
=
\phi_p^{\mathrm{per}}
\phi_p^{\mathrm{rrd}}
\Delta_p.
\label{eq:final_uncertainty}
\end{equation}
Therefore, the final path-equivalent rainfall result after realism-insufficiency correction and perception-consistency correction is expressed as
\begin{equation}
\hat{R}_{\mathrm{eq},p}
\pm
\Delta_{\mathrm{final},p}.
\label{eq:final_equivalent_result}
\end{equation}
Here, $\Delta_{\mathrm{final},p}$ integrates three sources of uncertainty: within-path rainfall-intensity fluctuation, insufficient rainfall realism, and additional risk caused by perception responses deviating from the real-rainfall reference range.

\section{Data Basis and Experimental Setup}

To support the credibility evaluation of simulated rainfall, this paper uses two types of real-rainfall reference data and conducts path sampling and perception experiments in a closed-field simulated-rainfall environment. The real-rainfall raindrop-distribution reference library is used to establish the physical distribution reference of natural rainfall, while the perception dataset provides reference perception responses under real-rainfall conditions. The closed-field experimental data serve as the evaluation object and are used to calculate the path-equivalent rainfall intensity, uncertainty band, and Realism of Raindrop Distribution (RRD) score, thereby supporting the selection of usable perception test paths.

\subsection{Real-Rainfall Reference Data}

The real-rainfall reference data used in this paper include a real-rainfall raindrop-distribution reference library and the RainSense real-rainfall perception dataset \cite{RainSense}. These two datasets serve different levels of the evaluation framework: the former describes the physical raindrop distribution of natural rainfall, whereas the latter characterizes the influence of real rainfall on autonomous-driving perception systems.

\begin{figure}[!t]
    \centering
    \includegraphics[width=0.9\linewidth]{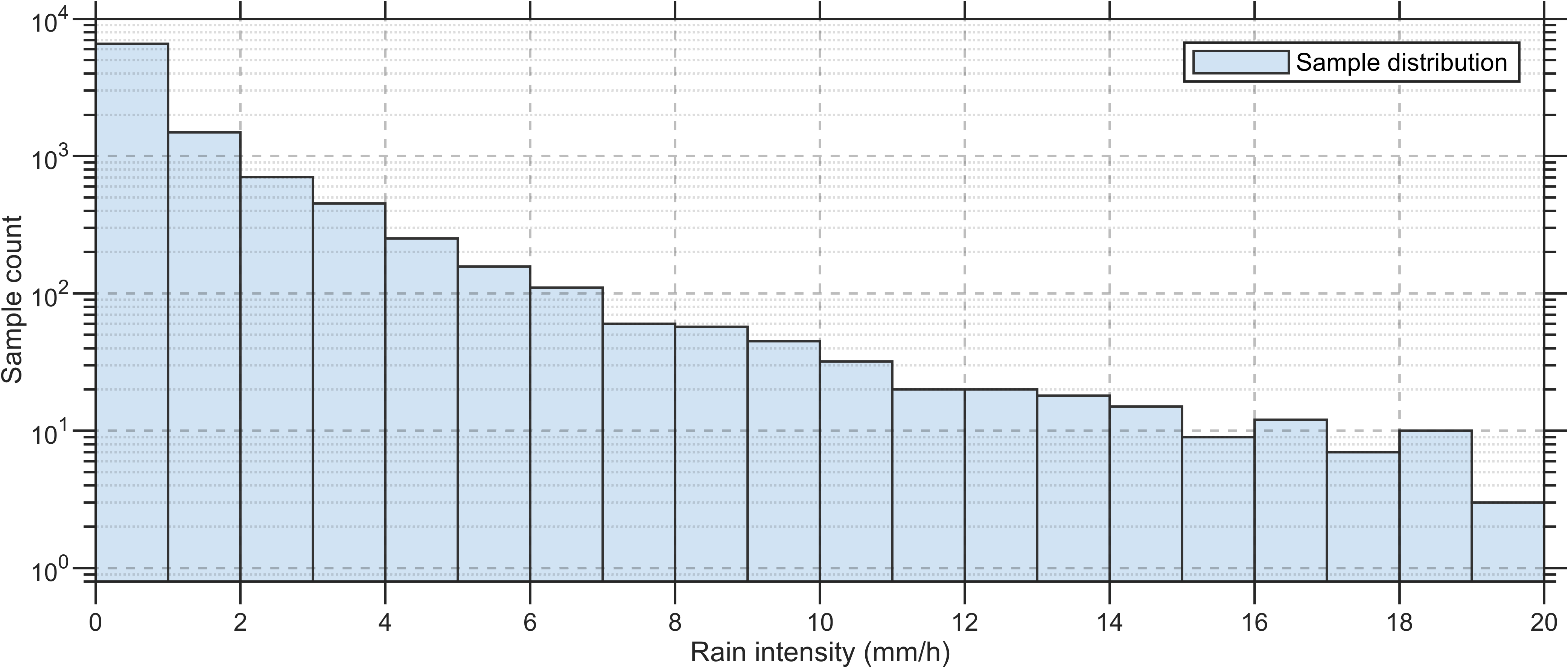}
    \caption{Rainfall-intensity distribution of the real-rainfall reference library.}
    \label{fig:real_rainfall_distribution}
\end{figure}

The real-rainfall raindrop-distribution reference library was collected in Zhengzhou, China. The data were acquired using a Laser-Optical Disdrometer deployed in an outdoor natural environment over a continuous one-year period. In total, more than 10,000 real-rainfall samples were collected, with each sample corresponding to a 1-min observation window and covering a rainfall-intensity range of 0--20~mm/h. Each sample contains the drop size--velocity distribution matrix and the corresponding rainfall intensity within the observation window. This reference library is used to construct raindrop-spectrum references under natural rainfall conditions, rather than as a single fixed template. In the evaluation process, real-rainfall samples with rainfall intensities close to the local intensity of each simulated-rainfall path point are selected from the library to form a dynamic real-rainfall baseline. Fig.~\ref{fig:real_rainfall_distribution} shows the rainfall-intensity distribution of the reference-library samples.

The RainSense dataset is used to provide reference perception responses under real-rainfall conditions. It contains 728 groups of multi-sensor perception data collected in real-rainfall environments, with rainfall intensities also covering 0--20~mm/h. In all collection scenes, dummy targets were placed at distances of 20~m and 40~m, and synchronized data were collected from lidar, camera, radar, and other onboard sensors. Since the posterior correction in this paper mainly focuses on the influence of simulated rainfall on lidar target-observation quality, the lidar point-cloud data corresponding to the 20~m dummy target in RainSense are selected. The target-surface point-cloud count and point-cloud reflectivity are extracted as reference perception indicators under real-rainfall conditions. Therefore, the real-rainfall raindrop-distribution reference library is used to determine whether the simulated raindrop distribution is close to natural rainfall, while RainSense is used to determine whether the perception response induced by simulated rainfall is consistent with that under real rainfall.

\begin{figure}[!t]
    \centering
    \includegraphics[width=0.9\linewidth]{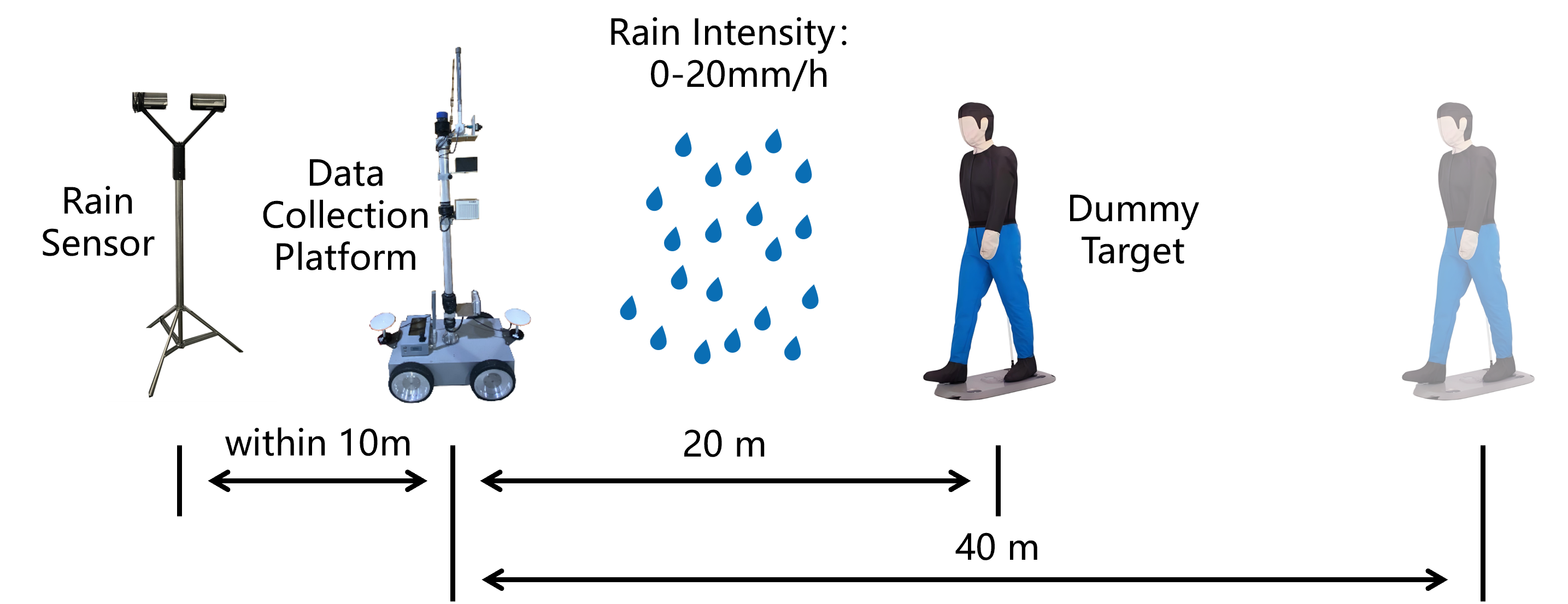}
    \caption{Data-collection setup of the RainSense dataset  \cite{RainSense}.}
    \label{fig:rainsense_setup}
\end{figure}

\subsection{Simulated-Rainfall Test Field and Sampling Layout}

\begin{figure*}[!t]
    \centering
    \includegraphics[width=0.88\textwidth]{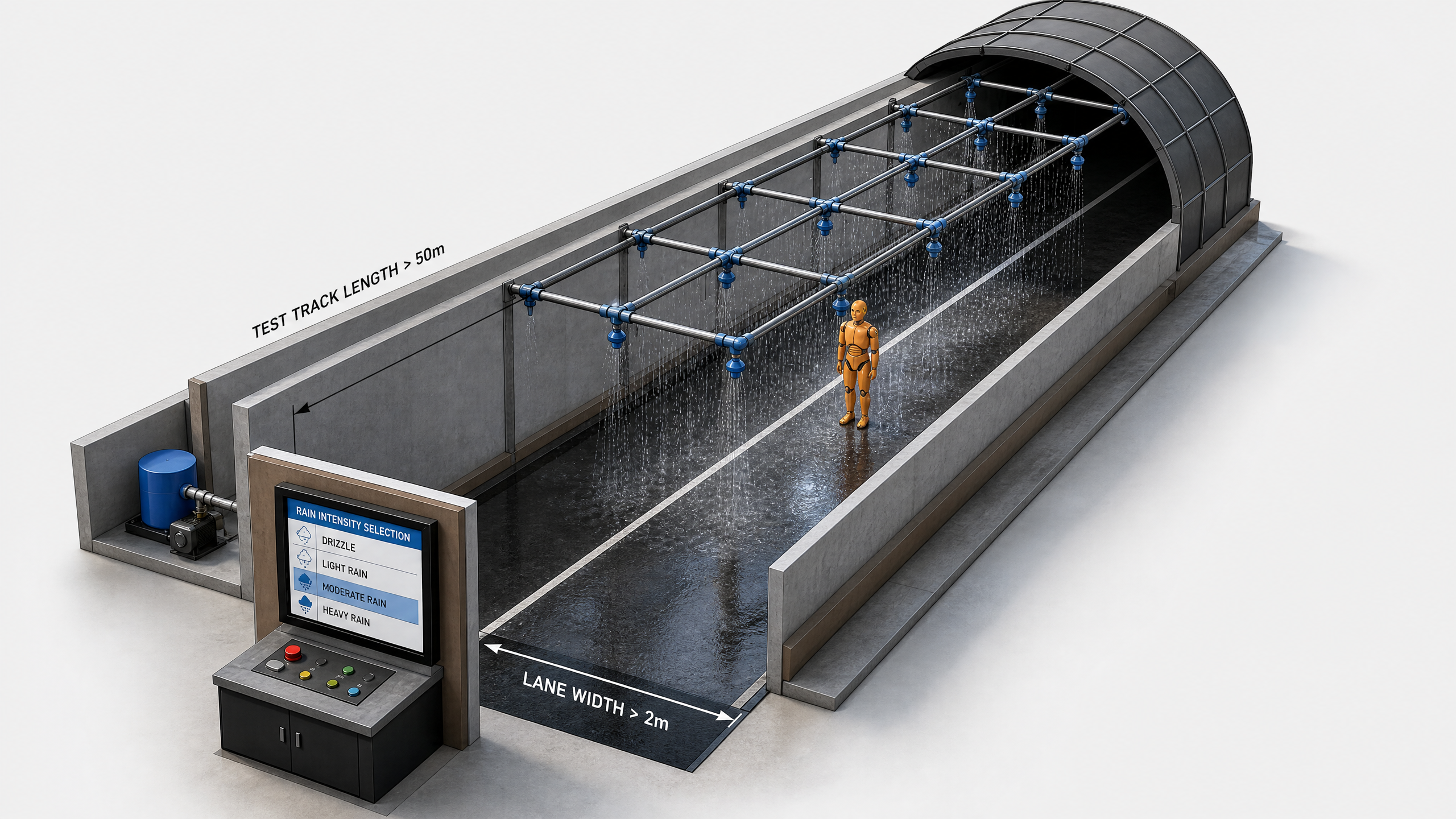}
    \caption{Schematic illustration of the closed-field simulated-rainfall test facility, showing the single-lane test track, overhead spray-nozzle array, and rainfall-intensity control unit.}
    \label{fig:test_field}
\end{figure*}

The simulated-rainfall experiments were conducted in a closed-field controllable rainfall test facility. The test area is a narrow single-lane space equipped with an overhead spray system, which can generate simulated rainfall within the lane region. The system supports multiple rainfall-intensity settings, including drizzle, light rain, moderate rain, and heavy rain, to cover different test requirements. In this study, the nominal drizzle condition is selected as the evaluation object. For de-identification, a schematic illustration is used in place of on-site photographs. As shown in Fig.~\ref{fig:test_field}, the facility consists of a single-lane test track, an overhead nozzle array, and a rainfall-intensity control unit, which together provide the basic infrastructure for closed-field simulated-rainfall experiments.

\begin{figure}[!t]
    \centering
    \includegraphics[width=0.9\linewidth]{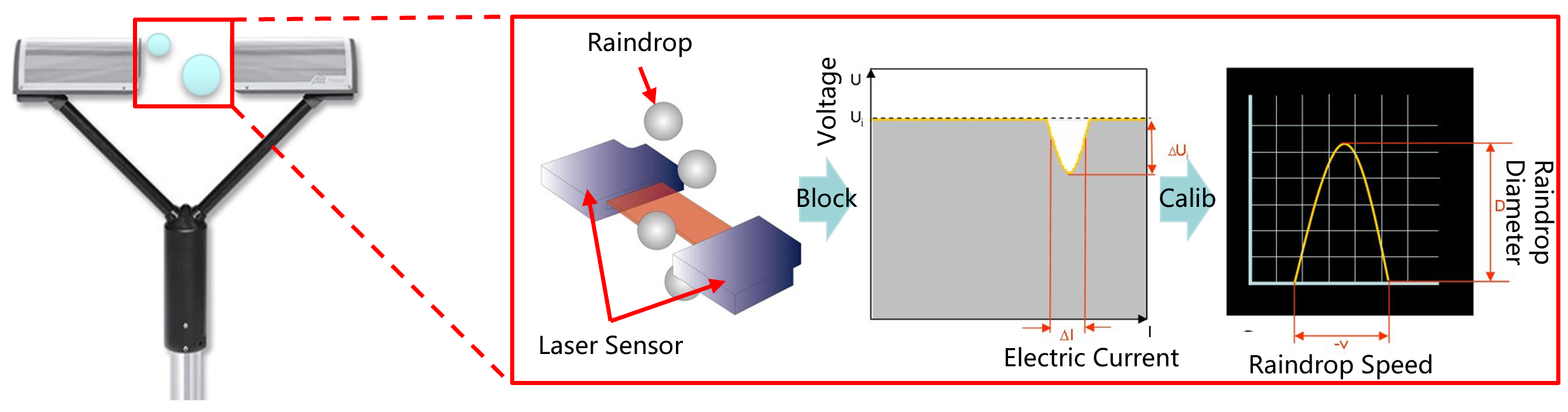}
    \caption{Measurement principle of the Laser-Optical Disdrometer.}
    \label{fig:disdrometer_principle}
\end{figure}

To obtain high-resolution spatial distribution data of the rain field, this study focuses on the minimum nozzle-group unit for refined measurement. This unit consists of the rainfall space corresponding to three nozzles and can be regarded as the basic operating unit of the simulated-rainfall system. Within this area, a two-dimensional sampling matrix is arranged, with a lateral width of 2.4~m and a longitudinal length of 7.2~m. Five sampling points are arranged laterally and nine sampling points are arranged longitudinally, resulting in 45 spatial sampling points under each nominal rainfall level. At each sampling point, the local rainfall intensity and drop size--velocity distribution matrix are collected to characterize the spatial distribution of simulated rainfall within the nozzle-group unit. Specifically, rainfall intensity and drop size--velocity distribution data are measured using a Parsivel2 Laser-Optical Disdrometer. The instrument measures raindrop diameter, fall velocity, and particle count based on the laser attenuation caused by precipitation particles. The measurement principle is illustrated in Fig.~\ref{fig:disdrometer_principle}.

\begin{figure}[!t]
    \centering
    \includegraphics[width=0.9\linewidth]{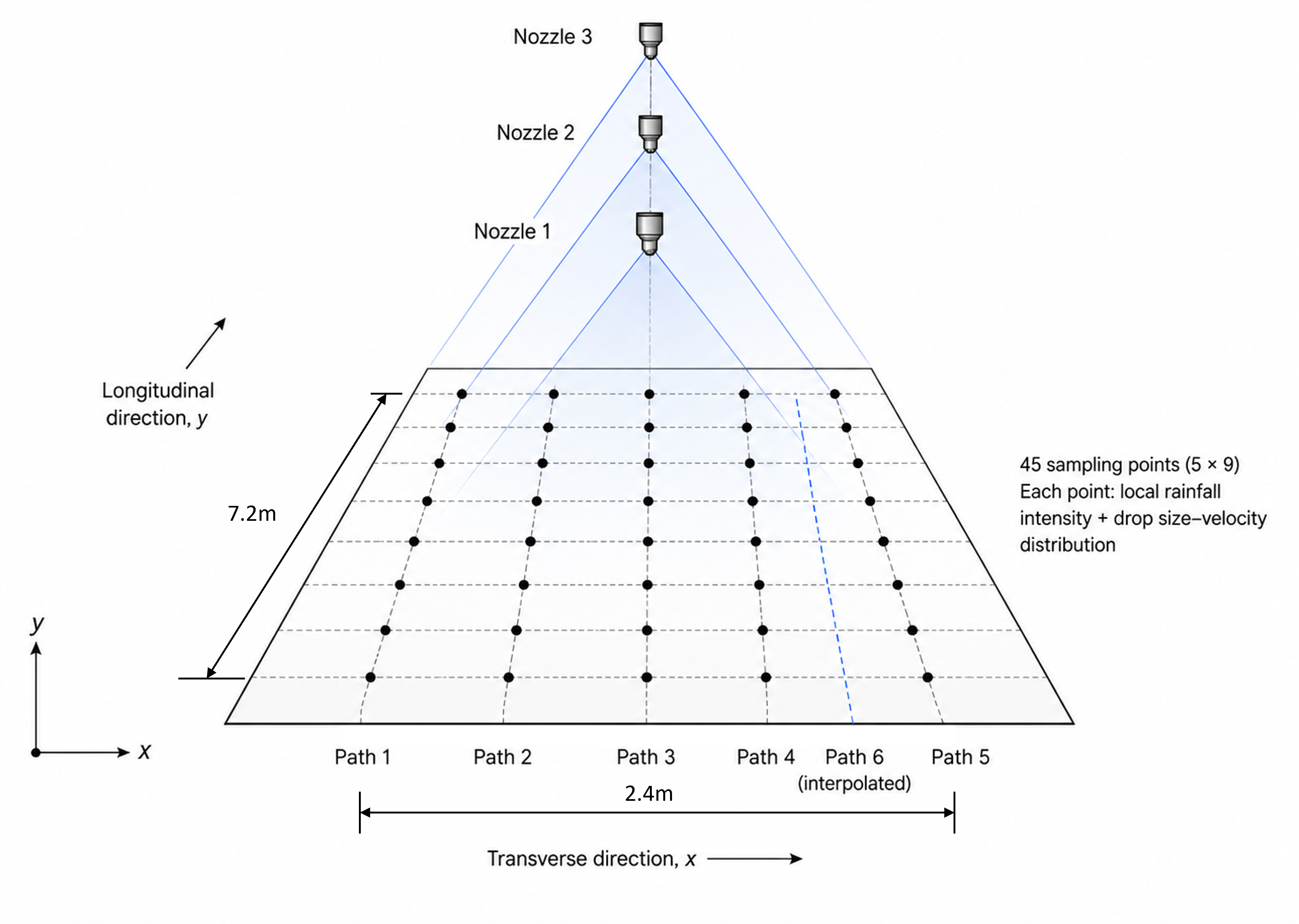}
    \caption{Spatial sampling layout of the simulated rain field.}
    \label{fig:sampling_layout}
\end{figure}

The test paths are arranged along the longitudinal direction of the lane. According to the five lateral sampling positions, five main candidate paths are formed and denoted as Path~I to Path~V. For intermediate positions between adjacent paths, the corresponding rainfall intensity and drop size--velocity distribution matrix are obtained by lateral interpolation under the assumption of spatial continuity. Based on the initial evaluation results, an additional Path~VI is further placed between Path~IV and Path~V as a candidate path for subsequent perception experiments. Therefore, the sampling matrix not only describes the spatial distribution of the rain field, but also provides unified spatial inputs for path-equivalent rainfall intensity calculation, RRD-score evaluation, and posterior perception-consistency correction. The overall sampling layout is shown in Fig.~\ref{fig:sampling_layout}.

\subsection{Posterior Perception Experiment and Indicator Extraction}

After rain-field sampling and preliminary evaluation of the candidate paths, dummy targets are placed on selected paths, and lidar point-cloud data are collected under simulated-rainfall conditions to obtain the observation indicators required for posterior perception-consistency correction. To improve the comparability between the simulated-rainfall experiment and the RainSense real-rainfall reference, the posterior perception experiment uses the same sensor and target configuration as RainSense: a Livox Tele-15 lidar and a standard dummy target. The dummy target at 20~m is selected as the main evaluation object, consistent with the corresponding real-rainfall observations in RainSense. Target-surface point clouds are extracted using the target region or bounding box, and the target-surface point-cloud count and mean point-cloud reflectivity are then calculated. These two indicators are compared with the real-rainfall perception reference range under similar rainfall intensities in RainSense, providing perception-level evidence for subsequent correction of the equivalent-rainfall uncertainty band.

In terms of indicator meaning, the target-surface point-cloud count mainly reflects the effective geometric sampling of the dummy target by lidar, and can characterize changes in target observability under rainfall. The mean point-cloud reflectivity reflects changes in the echo strength from the target surface, and can characterize the influence of raindrop scattering, occlusion, and attenuation on echo quality. These two indicators describe the influence of rainfall on lidar observation quality from the perspectives of target geometric completeness and echo-energy stability, respectively, and are therefore suitable as inputs for posterior perception-consistency correction.

The indicators are calculated as follows:
\begin{equation}
\bar{n}_{\mathrm{pc}}
=
\frac{n_{\mathrm{sum}}}{m},
\label{eq:mean_point_count}
\end{equation}
\begin{equation}
\bar{\rho}
=
\frac{
\sum_{i=1}^{m}
\sum_{j=1}^{n_i}
\rho_i^j
}{
n_{\mathrm{sum}}
}.
\label{eq:mean_reflectivity}
\end{equation}
Here, $\bar{n}_{\mathrm{pc}}$ denotes the mean point-cloud count, $n_{\mathrm{sum}}$ denotes the total number of target-region points across all frames within a given time interval, and $m$ denotes the number of collected frames. $\bar{\rho}$ denotes the mean reflectivity, $n_i$ is the number of points in the target region of the $i$-th frame, and $\rho_i^j$ is the reflectivity of the $j$-th point in the $i$-th frame.

It should be noted that the proposed posterior perception-consistency correction framework is not limited to lidar indicators. Depending on the test objective, it can also be extended to camera, radar, or fused-perception indicators. For example, target sharpness, contrast, and detection confidence in camera images, as well as target track count, echo strength, and detection stability in radar data, can be used as candidate posterior indicators. In this paper, lidar target point-cloud count and mean point-cloud reflectivity are selected because they directly reflect the influence of rainfall on three-dimensional target observation and can be stably extracted from the dummy-target region. This makes it possible to establish correspondence with the specific test path, target position, and real-rainfall reference in RainSense. In comparison, camera-based indicators are more easily affected by illumination, exposure, background texture, and image-processing strategies, whereas radar is less sensitive to rainfall and usually produces sparse target tracks, making it difficult to form stable posterior constraints at the current path scale.

\section{Results and Validation}

\subsection{Simulated Rain-Field Characteristic Analysis}

To evaluate the credibility of the simulated-rainfall test field, its overall characteristics are first analyzed from the perspective of microphysical distribution. Instead of relying only on nominal rainfall intensity or a few measurement points, this study introduces the drop size--velocity joint distribution to compare the microphysical morphology of simulated rainfall and real rainfall. This analysis reveals spatial differences in rainfall patterns within the test field and provides a basis for interpreting the subsequent path-based evaluation results.

\begin{figure*}[!t]
    \centering
    \includegraphics[width=0.95\textwidth]{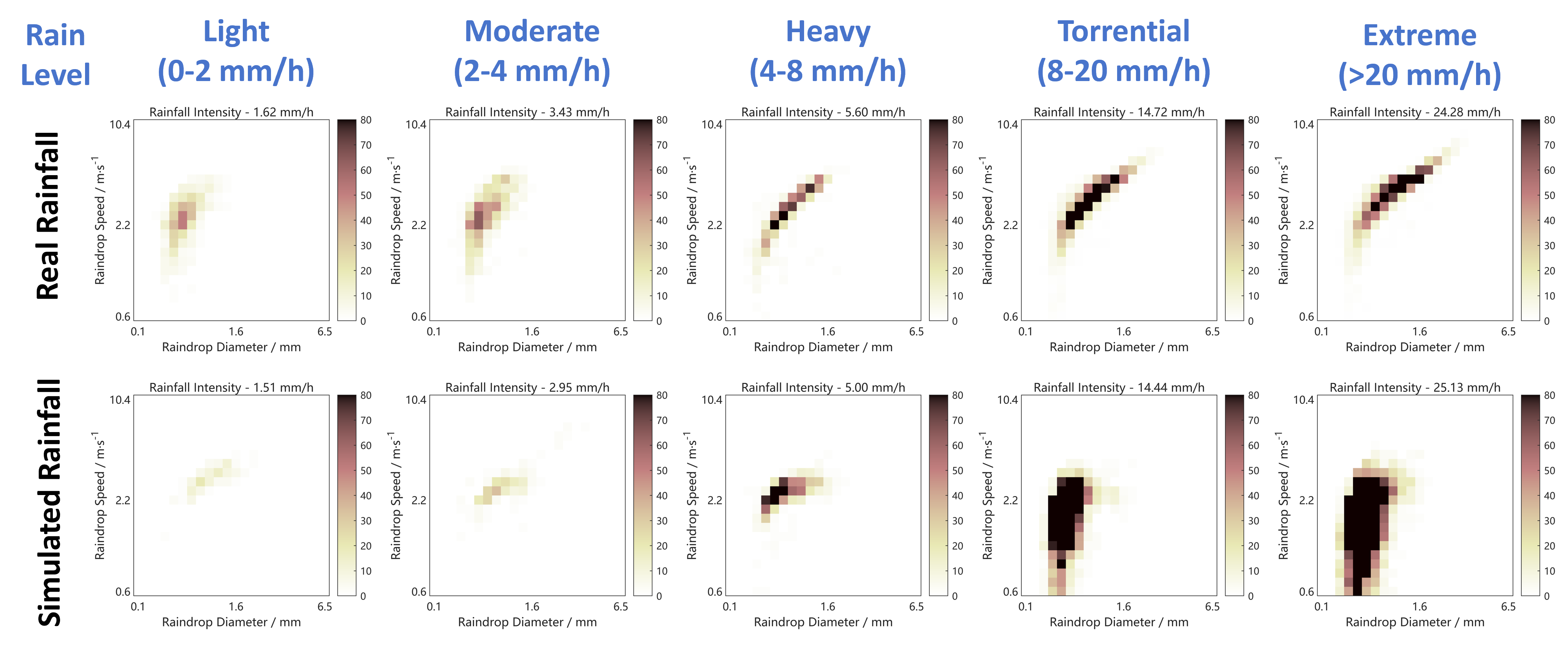}
    \caption{Comparison of drop size--velocity distributions between real rainfall and simulated rainfall under different rainfall-intensity levels.}
    \label{fig:real_simulated_micro_distribution}
\end{figure*}

As shown in Fig.~\ref{fig:real_simulated_micro_distribution}, clear differences exist between real rainfall and simulated rainfall under different rainfall-intensity levels. In real rainfall, raindrop diameter and fall velocity generally show a continuous correspondence, and the distribution expands toward larger diameters and higher velocities as rainfall intensity increases. In contrast, simulated rainfall contains a considerable concentration of low-velocity and small-diameter drops, even when the total number of drops increases under higher intensity levels. This indicates that a simulated-rainfall facility may generate a rainfall intensity close to the target level, but its microphysical raindrop distribution is not necessarily consistent with real rainfall. Therefore, realism evaluation at the level of the drop size--velocity joint distribution is necessary.

Furthermore, clustering analysis is performed on the drop size--velocity distribution matrices of the collected simulated-rainfall samples. Specifically, the raindrop counts in all diameter and velocity bins are used as features, and each sample is represented as a $32 \times 32 = 1024$-dimensional vector. K-means clustering is then used to identify typical microphysical distribution patterns. The results show that the simulated-rainfall samples can be mainly grouped into two representative patterns. One pattern shows a clear correlation between drop diameter and fall velocity, forming a bridge-shaped distribution. The other contains a high proportion of low-velocity and small-diameter drops, with a weak diameter--velocity correspondence, forming a column-shaped distribution.

\begin{figure}[!t]
    \centering
    \subfloat[Column-shaped distribution.]{
        \includegraphics[width=0.47\linewidth]{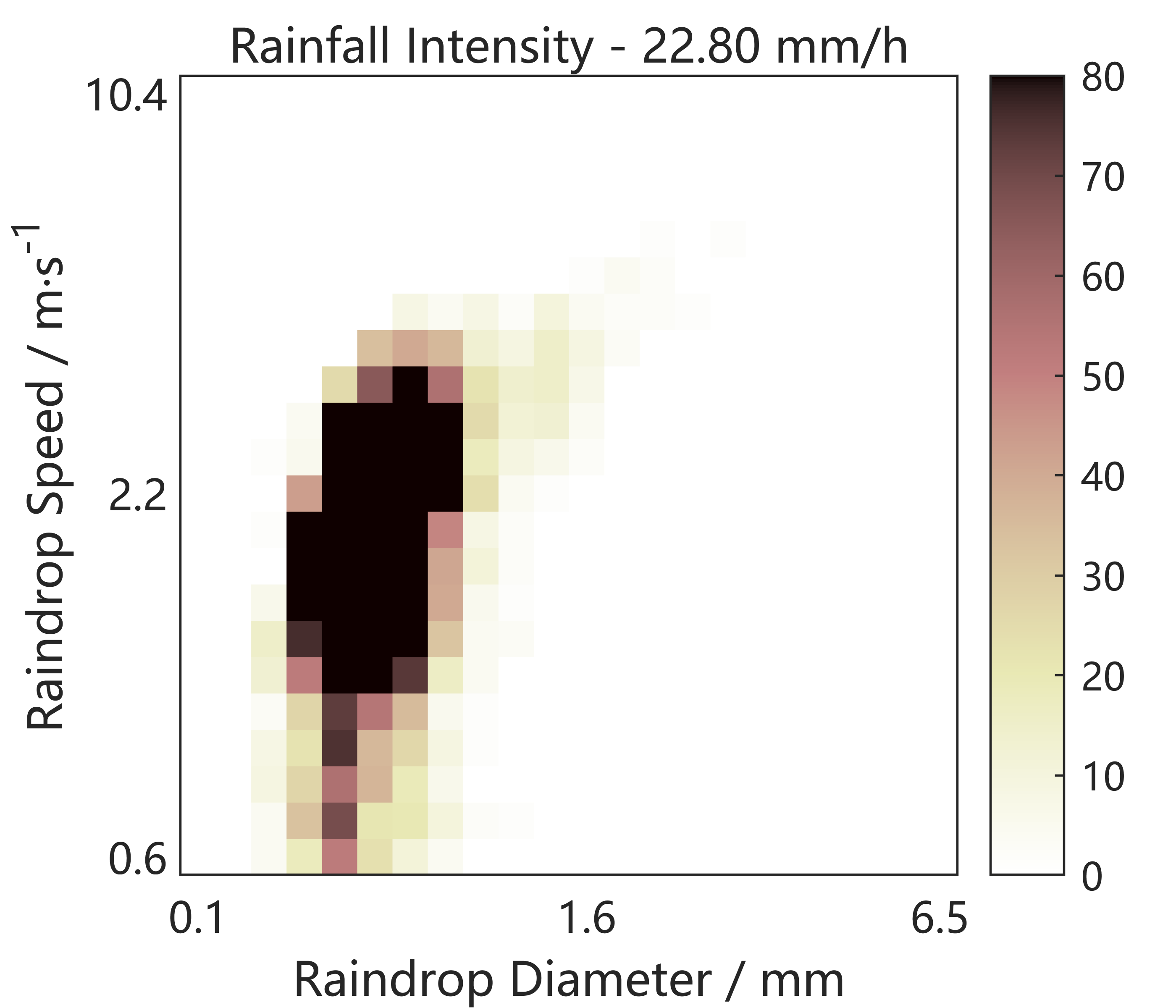}
        \label{fig:column_distribution}
    }
    \hfill
    \subfloat[Bridge-shaped distribution.]{
        \includegraphics[width=0.47\linewidth]{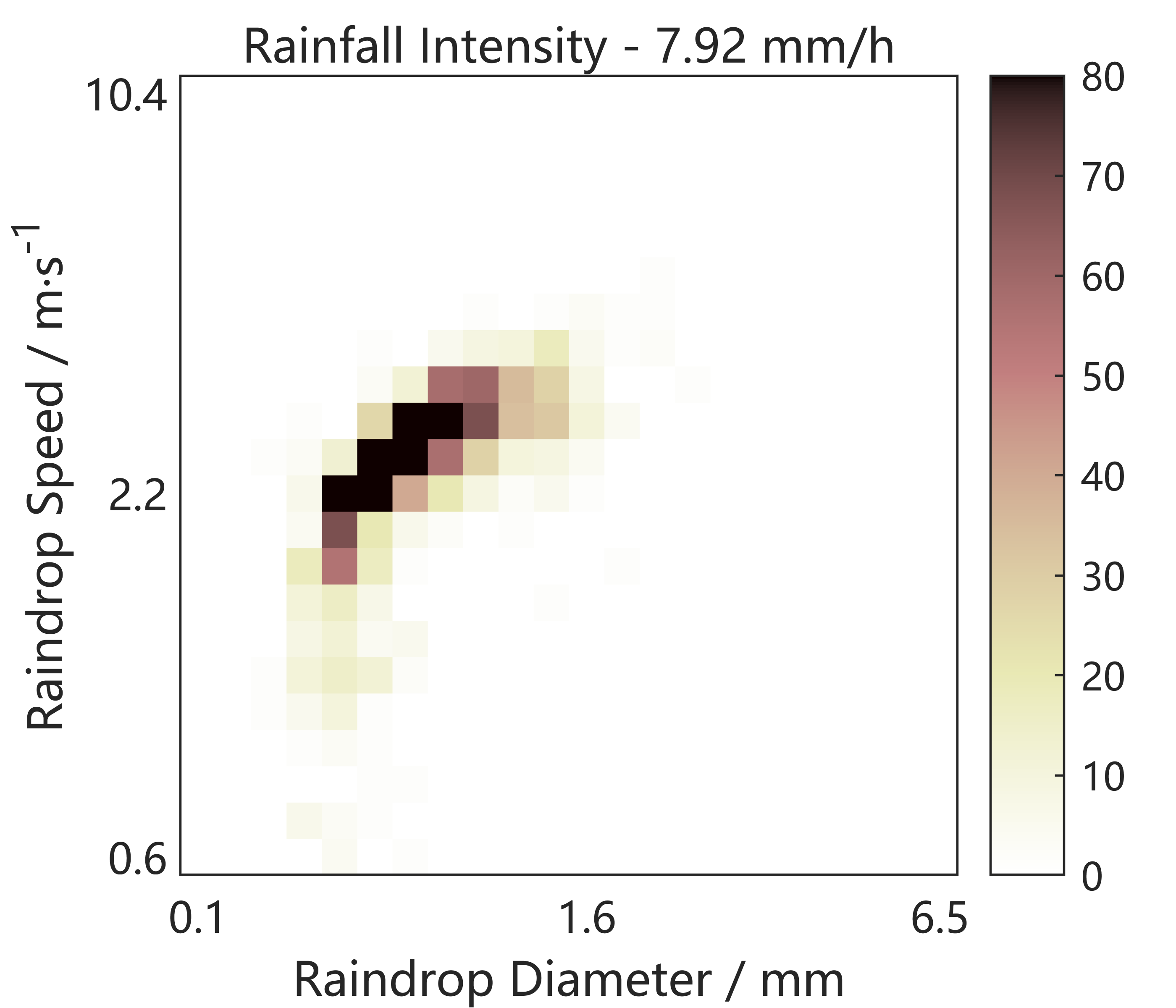}
        \label{fig:bridge_distribution}
    }
    \caption{Typical microphysical distribution patterns of simulated rainfall obtained by clustering.}
    \label{fig:simulated_rainfall_clustering}
\end{figure}

As shown in Fig.~\ref{fig:simulated_rainfall_clustering}, the bridge-shaped distribution is closer to real rainfall in overall morphology. As raindrop diameter increases, fall velocity also increases, which is consistent with the basic diameter--velocity relationship in natural rainfall. By contrast, the column-shaped distribution contains many low-velocity and small-diameter drops with concentrated velocities, indicating a clear discrepancy in raindrop dynamics from real rainfall. This pattern may be related to strong atomization in the spray system, droplet breakup producing many small drops, or the mismatch between local nozzle coverage and droplet motion. These observations show that although the simulated-rainfall system can generate a certain range of rainfall intensities, its microphysical distribution is not inherently equivalent to real rainfall.

\begin{figure*}[!t]
    \centering
    \includegraphics[width=0.95\textwidth]{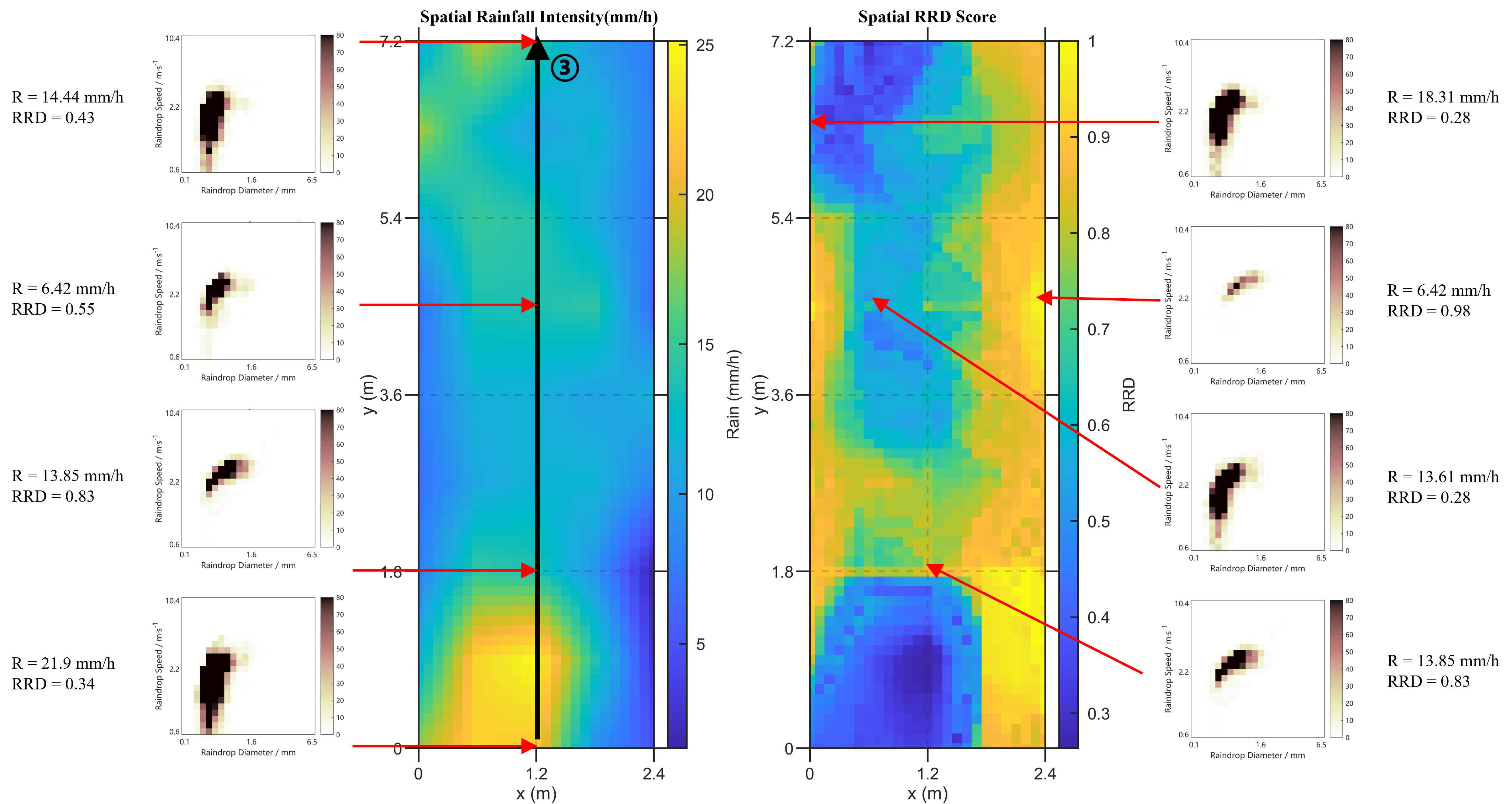}
    \caption{Spatial rainfall intensity, point-wise RRD, and representative microphysical distributions in the simulated rain field.}
    \label{fig:spatial_rainfield_rrd}
\end{figure*}

Based on the above analysis, the point-wise RRD is further calculated at each sampling position and visualized together with the spatial rainfall-intensity distribution. As shown in Fig.~\ref{fig:spatial_rainfield_rrd}, the spatial distribution of the simulated rain field is not random, but exhibits clear structural patterns. The rainfall-intensity map shows concentrated high- and low-intensity regions, indicating that the rainfall coverage generated by the nozzle group is not fully uniform. The point-wise RRD map further shows that the realism of the drop size--velocity distribution also has distinct spatial partitions. More importantly, the spatial structure of rainfall intensity does not simply coincide with the spatial structure of RRD. This suggests that spatial differences in the simulated rain field are not only caused by the amount of water, but are also related to nozzle-overlap effects, droplet breakup, atomization, and drop-fall states. Therefore, a spatial intensity map alone can describe the strength distribution of the field, but cannot determine whether the local raindrop-spectrum morphology is realistic.

The representative microphysical heatmaps at selected sampling points further show that the drop size--velocity distribution varies significantly across space. Some points exhibit a clear diameter--velocity correspondence, whereas others show concentrated low-velocity small drops or discontinuous distribution bands. This indicates that variations in point-wise RRD are not simply caused by rainfall-intensity changes, but reflect spatial differences in the microphysical formation process of simulated rainfall. Even along the same longitudinal path, the raindrop-spectrum morphology may vary considerably, implying that a vehicle experiences a continuously changing rainfall exposure process rather than a static condition represented by a single measurement point.

Overall, the simulated-rainfall field is not microphysically uniform. It contains both bridge-shaped samples that are closer to real rainfall and column-shaped samples with lower realism. Spatially, high-intensity regions, high-RRD regions, and low-credibility regions are interlaced. Therefore, the key of subsequent path-based evaluation is not simply to judge whether the simulated rain field is uniform, but to jointly consider rainfall intensity, microphysical realism, and spatial variation along each path under a real-rainfall reference, so as to identify candidate paths more suitable for perception testing.

\subsection{Path-Based Evaluation Results}

Based on the above characterization of the simulated rain field, candidate paths at different lateral positions in the test area are further evaluated. Since spatial non-uniformity remains under the same nominal simulated-rainfall condition, the purpose of path-based evaluation is to identify usable test paths with higher realism, better stability, and clearer rainfall-intensity representation from a non-uniform rain field, rather than merely judging the overall field uniformity.

\subsubsection{Baseline Evaluation Method}

To provide a comparison with the proposed path-based credibility evaluation, a conventional baseline is first constructed using path-wise rainfall statistics and uniformity indicators. Specifically, the longitudinal sampling points on each candidate path are treated as a rainfall-intensity sequence, and the path mean rainfall intensity $\bar{R}_p$, coefficient of variation $CV_p$, Christiansen uniformity coefficient $CU_p$, and low-quarter distribution uniformity $DU_{lq,p}$ are calculated. Here, $\bar{R}_p$ describes the average rainfall intensity corresponding to the path and is not used as a direct quality indicator. $CV_p$ measures the relative fluctuation of rainfall intensity within the path, where a smaller value indicates better stability. $CU_p$ describes the overall uniformity of the rainfall-intensity distribution, where a larger value indicates better spatial uniformity. $DU_{lq,p}$ further reflects the adequacy of low-intensity coverage, where a larger value indicates fewer low-intensity weak regions. These indicators are all calculated from rainfall intensity and reflect the traditional focus on intensity level, fluctuation, and spatial uniformity in simulated-rainfall evaluation \cite{24,27,28}.

\begin{equation}
\bar{R}_p =
\frac{1}{n}
\sum_{i=1}^{n}
R_{p,i},
\label{eq:path_mean_rainfall}
\end{equation}

\begin{equation}
CV_p =
\frac{\sigma_{R,p}}{\bar{R}_p},
\label{eq:path_cv}
\end{equation}

\begin{equation}
CU_p =
1 -
\frac{
\sum_{i=1}^{n}
\left|
R_{p,i} - \bar{R}_p
\right|
}{
n\bar{R}_p
},
\label{eq:path_cu}
\end{equation}

\begin{equation}
DU_{lq,p} =
\frac{
\bar{R}_{\mathrm{low25\%},p}
}{
\bar{R}_p
}.
\label{eq:path_du_lq}
\end{equation}

Here, $R_{p,i}$ denotes the local rainfall intensity at the $i$-th longitudinal sampling point of path $p$, $n$ is the number of sampling points on the path, $\sigma_{R,p}$ is the standard deviation of the rainfall-intensity sequence, and $\bar{R}_{\mathrm{low25\%},p}$ is the mean rainfall intensity of the lowest 25\% sampling points.

The baseline evaluation results are shown in Table~\ref{tab:baseline_path_evaluation}. Different candidate paths exhibit clear differences in mean rainfall intensity, intensity fluctuation, and uniformity. Path~II and Path~III have relatively high mean rainfall intensities of 15.841~mm/h and 15.183~mm/h, respectively, but their $CV_p$ values are higher than those of Path~IV and Path~VI, and their $CU_p$ and $DU_{lq,p}$ values are not optimal. This indicates that higher rainfall intensity does not necessarily imply better path uniformity. Path~I and Path~V show poorer uniformity, especially Path~V, which has the highest $CV_p$ and the lowest $CU_p$ and $DU_{lq,p}$ among all paths. By contrast, Path~IV and Path~VI show more stable performance under conventional uniformity indicators. Path~IV has the lowest $CV_p$ and the highest $CU_p$ and $DU_{lq,p}$, while Path~VI also has a low $CV_p$ and a $CU_p$ close to that of Path~IV. Therefore, if only rainfall-intensity fluctuation and spatial uniformity are considered, Path~IV and Path~VI can be regarded as preferable candidate paths. The following analysis further introduces drop size--velocity distribution realism and corrected uncertainty bands to compare with this conventional screening result.

\begin{table}[!t]
\centering
\caption{Baseline Path Evaluation Results}
\label{tab:baseline_path_evaluation}
\renewcommand{\arraystretch}{1.12}
\setlength{\tabcolsep}{4pt}
\begin{tabular}{lcccc}
\hline
Path & $\bar{R}_p$ & $CV_p \downarrow$ & $CU_p \uparrow$ & $DU_{lq,p} \uparrow$ \\
\hline
Path~I   & 10.956 & 0.410 & 0.687 & 0.644 \\
Path~II  & 15.841 & 0.287 & 0.765 & 0.740 \\
Path~III & 15.183 & 0.329 & 0.756 & 0.735 \\
Path~IV  & 11.305 & 0.168 & 0.865 & 0.831 \\
Path~V   & 6.414  & 0.458 & 0.661 & 0.492 \\
Path~VI  & 8.860  & 0.196 & 0.850 & 0.785 \\
\hline
\end{tabular}
\end{table}

\subsubsection{Proposed Path-Based Evaluation}

After the preliminary screening based on conventional macroscopic uniformity indicators, the proposed path-based credibility evaluation method is used to analyze the real-rainfall proxy capability of the candidate paths. The baseline results indicate that Path~IV and Path~VI have favorable intensity stability and path continuity. However, conventional macroscopic indicators can only indicate whether rainfall intensity along a path is relatively uniform. They cannot determine whether the drop size--velocity joint distribution is close to real rainfall, nor can they represent the uncertainty in subsequent perception tests. Therefore, path-equivalent rainfall intensity, corrected uncertainty band, and path-averaged RRD score are further introduced to verify and extend the conventional screening result.

\begin{figure*}[!t]
    \centering
    \includegraphics[width=0.76\textwidth]{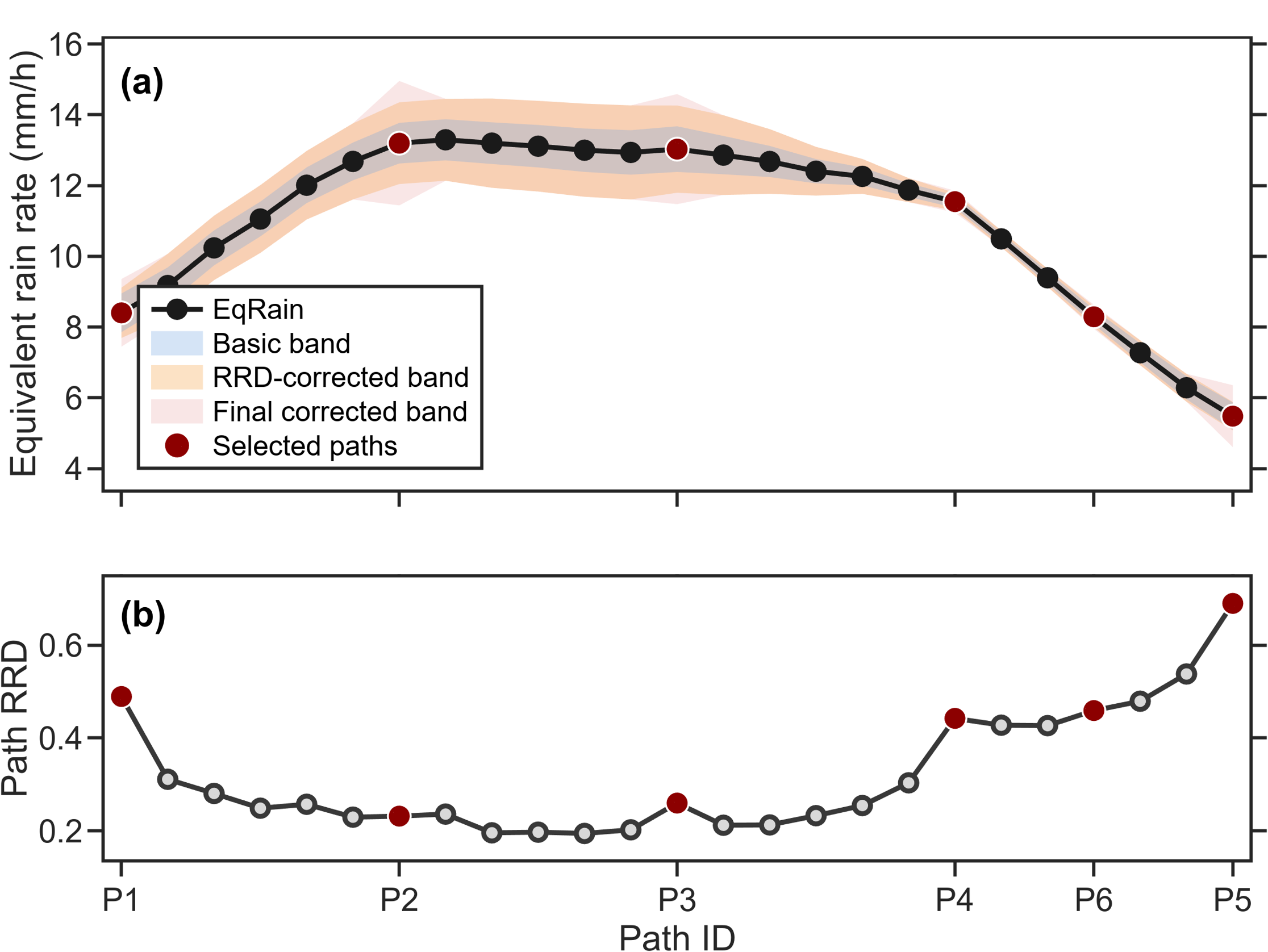}
    \caption{Path-based credibility evaluation results of the candidate paths.}
    \label{fig:path_evaluation_results}
\end{figure*}

Several representative longitudinal paths are selected in the sampling area, and their path-equivalent rainfall intensity, basic uncertainty band, corrected uncertainty band, and path-averaged RRD score are calculated. Fig.~\ref{fig:path_evaluation_results} shows the path locations, the distribution of path-equivalent rainfall intensity and uncertainty bands, and the trend of path-averaged RRD scores.

\begin{figure*}[!t]
    \centering
    \includegraphics[width=0.76\textwidth]{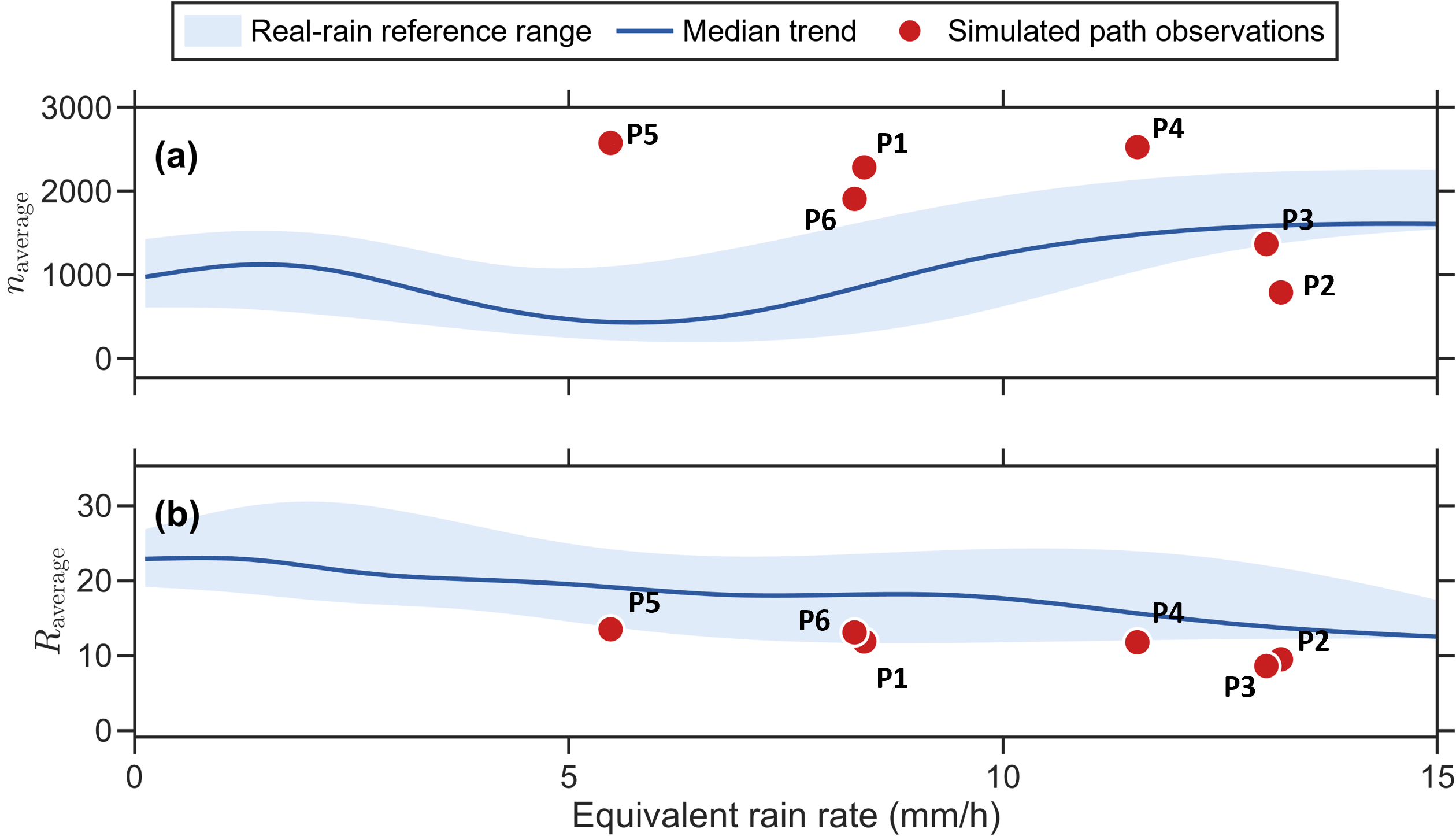}
    \caption{Perception-consistency results of the candidate paths under simulated rainfall.}
    \label{fig:perception_consistency}
\end{figure*}

Overall, the evaluation results differ significantly among paths, indicating that even within the same test field and under the same simulated-rainfall experiment, the rainfall conditions experienced along different paths are not identical. Consistent with the conventional macroscopic indicators, Path~IV and Path~VI are still identified as preferable candidate paths by the proposed method. They have relatively small corrected uncertainty bands and relatively high path-averaged RRD scores, indicating not only good rainfall-intensity stability but also better raindrop-spectrum realism and result interpretability. In addition, the posterior perception-consistency results in Fig.~\ref{fig:perception_consistency} show that Path~IV and Path~VI are closer to the RainSense real-rainfall reference ranges in terms of both $\bar{n}_{\mathrm{pc}}$ and $\bar{\rho}$. This suggests that they perform better not only at the rain-field evaluation level, but also in terms of real-rainfall proxy capability for lidar target observation.

By contrast, although Path~II and Path~III have higher equivalent rainfall intensities and are closer to the center of nozzle coverage, they have lower path-averaged RRD scores and wider corrected uncertainty bands. Their posterior perception indicators also lie near or below the lower side of the real-rainfall reference range, indicating deviations from real-rainfall perception responses at comparable rainfall intensities. Path~I and Path~V show another type of discrepancy. Path~V has a relatively high path-averaged RRD score, but its $\bar{n}_{\mathrm{pc}}$ is significantly higher than the real-rainfall reference range, which enlarges the uncertainty band after posterior correction. A similar high point-count phenomenon is also observed for Path~I. These results show that relying only on macroscopic uniformity or raindrop-spectrum realism may lead to incomplete judgments of path usability, while posterior perception consistency further reveals differences in path applicability for perception testing.

Therefore, the proposed method is consistent with the conventional method in its main screening conclusion, but provides richer information. Conventional indicators show that Path~IV and Path~VI are favorable in terms of rainfall-intensity uniformity and along-path stability. The proposed method further shows that these two paths also have higher RRD scores and smaller corrected uncertainty bands. In comparison, Path~II and Path~III exhibit high rainfall intensity but insufficient realism and uncertainty control, while Path~V shows high microphysical realism but increased uncertainty after posterior correction. Thus, the proposed method provides incremental information beyond conventional uniformity evaluation, making the advantages and limitations of different paths clearer and supporting more complete path selection for simulated-rainfall perception tests.

\begin{table*}[!t]
\centering
\caption{Ablation Comparison of Evaluation Schemes}
\label{tab:ablation_study}
\renewcommand{\arraystretch}{1.15}
\setlength{\tabcolsep}{4pt}
\footnotesize
\begin{tabular*}{\textwidth}{@{\extracolsep{\fill}}l p{2.3cm} p{3.0cm} p{3.0cm} p{3.8cm}@{}}
\hline
Item & Baseline~A & Baseline~B & Baseline~C & Proposed \\
\hline
Components 
& $\hat{R}_{\mathrm{eq},p}$ 
& $\hat{R}_{\mathrm{eq},p}$, $\overline{\mathrm{RRD}}_p$ 
& $\hat{R}_{\mathrm{eq},p}\pm\Delta_p$ 
& $\hat{R}_{\mathrm{eq},p}\pm\Delta_{\mathrm{final},p}$, $\overline{\mathrm{RRD}}_p$ \\

Path~I   
& 8.40 
& 8.40, 0.49 
& $8.40\pm0.54$ 
& $8.40\pm0.95$, 0.49 \\

Path~II  
& 13.19 
& 13.19, 0.23 
& $13.19\pm0.57$ 
& $13.19\pm1.75$, 0.23 \\

Path~III 
& 13.03 
& 13.03, 0.25 
& $13.03\pm0.64$ 
& $13.03\pm1.55$, 0.25 \\

Path~IV  
& 11.54 
& 11.54, 0.43 
& $11.54\pm0.18$ 
& $11.54\pm0.31$, 0.43 \\

Path~V   
& 5.48 
& 5.48, 0.69 
& $5.48\pm0.38$ 
& $5.48\pm0.87$, 0.69 \\

Path~VI  
& 8.28 
& 8.28, 0.46 
& $8.28\pm0.21$ 
& $8.28\pm0.34$, 0.46 \\

\hline
Preferred paths 
& II/III 
& I/V 
& IV/V/VI 
& IV/VI \\
\hline
\multicolumn{5}{@{}l}{\footnotesize Note: In Baseline~B and Proposed, the second value in each path cell denotes $\overline{\mathrm{RRD}}_p$.}
\end{tabular*}
\end{table*}

\subsection{Method Validation}

To verify the necessity of different components in the proposed method, four ablation schemes are designed for comparison. These schemes correspond to the progressive extension from conventional rainfall-intensity representation to the complete evaluation framework. Baseline~A uses only path-equivalent rainfall intensity for evaluation. Baseline~B introduces the path-averaged RRD score $\overline{\mathrm{RRD}}_p$ in addition to equivalent intensity, but does not use the corrected uncertainty band. Baseline~C uses path-equivalent rainfall intensity with the basic uncertainty band. The Proposed method outputs the path-equivalent rainfall intensity, corrected uncertainty band, and path-averaged RRD score simultaneously. By comparing the path rankings and recommendations under different schemes, the risk of misleading decisions caused by a single indicator or an incomplete framework can be analyzed.

As shown in Table~\ref{tab:ablation_study}, different ablation schemes lead to different path recommendations. Baseline~A selects Path~II and Path~III because it only considers the highest path-equivalent rainfall intensity, but it cannot identify their insufficient microphysical realism. Baseline~B shifts the recommendation to Path~I and Path~V after introducing $\overline{\mathrm{RRD}}_p$, indicating that RRD can partly correct the bias of using only rainfall intensity; however, it may still favor paths with high RRD scores without sufficiently constraining uncertainty. Baseline~C recommends Path~IV, Path~V, and Path~VI because it considers the basic uncertainty band and reflects within-path rainfall-intensity fluctuation, but it cannot determine whether the stability is built on similarity to real rainfall.

By contrast, the Proposed method outputs both $\hat{R}_{\mathrm{eq},p}\pm\Delta_{\mathrm{final},p}$ and $\overline{\mathrm{RRD}}_p$, and its recommendation shifts from extreme single-indicator paths to Path~IV and Path~VI. This shows that the complete framework does not simply pursue the highest rainfall intensity, the highest RRD score, or the smallest basic uncertainty band, but performs a joint screening under three constraints. In particular, Path~V is easily regarded as preferable in both Baseline~B and Baseline~C, but is no longer prioritized in the Proposed method because its final uncertainty band is enlarged. This indicates that the corrected uncertainty band can reveal potential application uncertainty. The posterior perception indicators are not discussed again as independent validation metrics to avoid circular reasoning; their role is mainly reflected in the effect of $\Delta_{\mathrm{final},p}$ on path ranking, where uncertainty correction prevents overly optimistic judgments for certain paths.

\section{Conclusion}

This paper proposes a real-rainfall-reference-based, path-based credibility evaluation framework for simulated rainfall. The framework establishes a spatial equivalence evaluation chain from raindrop-spectrum physical characteristics to perception-system responses, addressing the lack of quantitative path-based characterization methods for simulated rain fields in autonomous-driving perception tests. Unlike conventional methods that describe test conditions only by nominal or mean rainfall intensity, the proposed method represents each candidate path by a comprehensive result consisting of path-equivalent rainfall intensity, uncertainty band, and path-averaged Realism of Raindrop Distribution (RRD) score. This representation jointly characterizes the real-rainfall intensity represented by the path, the uncertainty of this interpretation, and the consistency of the drop size--velocity joint distribution with real rainfall. 

Validation based on approximately 10,000 real-rainfall raindrop-spectrum samples, 728 real-rainfall perception samples, and 45 spatial sampling points in the simulated rain field shows that clear spatial and microphysical differences exist within the simulated rain field, and that rainfall-intensity distribution is not fully consistent with microphysical realism. The path-based evaluation further indicates that Path~IV and Path~VI achieve more balanced performance in macroscopic stability, microphysical realism, and perception consistency. In contrast, some paths have relatively high rainfall intensity but insufficient microphysical realism, whereas others are closer to real rainfall in raindrop-spectrum structure but show larger deviations in perception response. The ablation results show that relying only on rainfall intensity, RRD score, or the basic uncertainty band may lead to biased path selection. By integrating these factors, the complete framework can more robustly identify test paths with rainfall-intensity compatibility, low uncertainty, and real-rainfall proxy capability, thereby providing a quantitative basis for path selection, condition description, and result interpretation in simulated-rainfall perception tests. 

Future work will extend the validation to more simulated-rainfall test fields, rainfall-intensity levels, and perception-task scenarios. This will further examine the applicability of the proposed method under different test facilities and sensor configurations, and provide methodological support for quantitative SOTIF evaluation of perception systems under rainfall and the standardization of related test conditions.

\end{document}